\documentclass{article}
\usepackage[preprint]{neurips_2024}

\usepackage[utf8]{inputenc} 
\usepackage[T1]{fontenc}             
\usepackage{booktabs}            
\usepackage{nicefrac}       
\usepackage{microtype}      
\usepackage{xspace}
\usepackage{algpseudocode}
\usepackage{algorithm}
\usepackage{wrapfig}
\usepackage{graphicx}
\usepackage{multirow}
\usepackage{tikz}
\usepackage{xcolor}
\usepackage{amsthm}
\usepackage{amssymb}
\usepackage{amsfonts}
\usepackage{amsmath}
\usepackage{bbm}
\usepackage{subcaption}
\usepackage[scaled=0.86]{helvet}

\usepackage{amsmath,amsfonts,bm}

\def\eqref#1{equation~\ref{#1}}

\def\1{\bm{1}}

\DeclareMathAlphabet{\mathsfit}{\encodingdefault}{\sfdefault}{m}{sl}
\SetMathAlphabet{\mathsfit}{bold}{\encodingdefault}{\sfdefault}{bx}{n}

\usepackage{enumitem}
\setlist[enumerate]{itemsep=0.2ex, topsep=0.2\topsep}
\setlist[description]{itemsep=0.2ex, topsep=0.2\topsep}
\setlist[itemize]{leftmargin=3ex,itemsep=0.2ex, topsep=0.2\topsep}

\theoremstyle{definition}
\newtheorem{example}{Example}[section]

\newtheorem{proposition}{Proposition}[section]
\newtheorem{theorem}{Theorem}[section]

\usepackage{hyperref}
\hypersetup{%
    colorlinks=true,
    linkcolor=darkgray,
    citecolor=darkgray,
   urlcolor  = gray}
\usepackage{url}
\PassOptionsToPackage{numbers}{natbib}

\newcommand{\var}{\mathsf{Var}}

\newcommand{\con}{\mathsf{Con}}
\newcommand{\pathend}{\mathsf{end}}
\newcommand{\pathvalid}{\mathsf{valid}}
\newcommand{\conocc}{\mathsf{ConOcc}}
\newcommand{\prob}{\mathbb{P}}

\newcommand{\G}{\mathcal{G}}

\newcommand{\E}{\mathcal{E}}
\newcommand{\cP}{\mathcal{P}}
\newcommand{\V}{\mathcal{V}}

\newcommand{\R}{\mathcal{R}}
\newcommand{\W}{\mathcal{W}}

\newcommand{\unrav}[2]{\textsf{UnRAvL}_{#2}(#1)}   
\newcommand{\bx}{\mathbf{x}}
\newcommand{\In}{\textsf{In}}
\newcommand{\Out}{\textsf{Out}}
\newcommand{\Neigb}{\textsf{N}}

\newcommand{\qta}{\textsf{UnRavL}\xspace}
\newcommand{\unr}{\textsf{UnRavL}\xspace}
\newcommand{\nbfnets}{\textsf{NBFNets}\xspace}
\newcommand{\nbfnet}{\textsf{NBFNet}\xspace}
\newcommand{\baseline}{\textsf{PQE}\xspace}
\newcommand{\mpqe}{\textsf{MPQE}\xspace}
\newcommand{\gnnqe}{\textsf{GNN-QE}\xspace}
\let\mc\multicolumn

\newcommand*{\eox}[1][{\textcolor{darkgray!70}{{\large $\diamond$}}}]{%
\leavevmode\unskip\penalty9999 \hbox{}\nobreak\hfill
    \quad\hbox{#1}%
}
\newcommand*{\eop}[1][{\textcolor{darkgray!70}{$\square$}}]{%
\leavevmode\unskip\penalty9999 \hbox{}\nobreak\hfill
    \quad\hbox{#1}%
}

\usepackage{tikz-cd}\newcommand*\circled[1]{\tikz[baseline=(char.base)]{
            \node[shape=circle,draw,inner sep=0pt] (char) {#1};}}

\title{A Neuro-Symbolic Framework for Answering Graph Pattern Queries in Knowledge Graphs}

\author{
  Tamara Cucumides$^{1}$ \quad
 Daniel Daza$^{2,3}$\quad 
  Pablo Barcel{\'o}$^{4,5}$  \\
  \textbf{Michael Cochez$^{2,3}$}\quad 
  \textbf{Floris Geerts$^{1}$} \quad
   \textbf{Miguel Romero$^{5,6}$} \quad 
   \textbf{Juan L Reutter$^{5,6}$}\\
$^1$Department of Computer Science, University of Antwerp\\
$^2$Computer Science, Vrije Universiteit Amsterdam\\
$^3$Discovery Lab, Elsevier, Amsterdam\\
 $^4$Inst. for Math. and Comp. Eng., Universidad Cat\'{o}lica de Chile \& IMFD Chile \\
$^5$CENIA Chile\\
$^6$Department of Computer Science, 
  Universidad Cat\'olica de Chile\\
  \texttt{\{tamara.cucumidesfaundez,floris.geerts\}@uantwerp.be}\\
  \texttt{\{pbarcelo,mgromero,jreutter\}@uc.cl}\\
\texttt{\{m.cochez,d.dazacruz\}@vu.nl}
}

\begin{document}

\maketitle

\begin{abstract}
The challenge of answering graph queries over incomplete knowledge graphs is gaining significant attention in the machine learning community. Neuro-symbolic models have emerged as a promising approach, combining good performance with high interpretability. These models utilize trained architectures to execute atomic queries and integrate modules that mimic symbolic query operators. However, most neuro-symbolic query processors are constrained to \emph{tree-like} graph pattern queries. These queries admit a bottom-up execution with constant values or \emph{anchors} at the leaves and the target variable at the root. While expressive, tree-like queries fail to capture critical properties in knowledge graphs, such as the existence of multiple edges between entities or the presence of triangles. 
We introduce a framework for answering arbitrary graph pattern queries over incomplete knowledge graphs, encompassing both cyclic queries and tree-like queries with existentially quantified leaves. These classes of queries are vital for practical applications but are beyond the scope of most current neuro-symbolic models. Our approach employs an approximation scheme that facilitates acyclic traversals for cyclic patterns, thereby embedding additional symbolic bias into the query execution process.
Our experimental evaluation demonstrates that our framework performs competitively on three datasets, effectively handling cyclic queries through our approximation strategy. Additionally, it maintains the performance of existing neuro-symbolic models on anchored tree-like queries and extends their capabilities to queries with existentially quantified variables.
\end{abstract}

\section{Introduction}\label{sec:intro}

Knowledge graphs are prevalent in both industry and the scientific community, playing a crucial role in representing organizational knowledge \citep{Fensel_2020,Hogan_2021}. They model information as nodes (entities) and edges (relations between entities). However, many applications using knowledge graphs encounter the issue of \emph{missing} information. As knowledge graphs are created or updated, their information may become stale or conflicting, and certain data sources might not be integrated yet. Consequently, knowledge graphs often remain \emph{incomplete}, lacking some entities or relations relevant to the application domain 
\citep{Ren-Survey_2023}.

A particularly important reasoning task on knowledge graphs is \emph{answering queries}. Traditional query answering methods, especially those from data management and semantic web literature, focus on symbolic queries specified in specific symbolic query languages, and on extracting the information that can be derived from the knowledge \emph{present} in the graph~\citep{gdbs-survey2017,Hogan_2021, survey-aidan-rdf-engines}. 
Given the incomplete nature of knowledge graphs, these methods fail to address the need to reason about unknown information, reducing their usefulness in many application domains \citep{nickel2015review}.

\begin{figure}
\centering
\includegraphics[height=3.4cm]{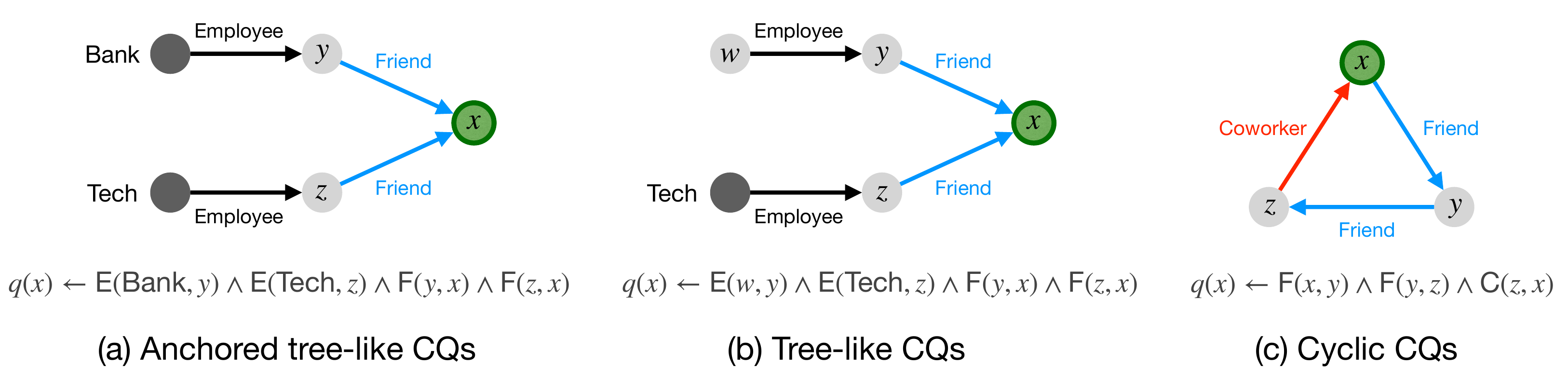}
\caption{(a) Edges in \emph{anchored tree-like} queries are structured as trees where the leaves are anchors and the root is the target variable (here $x$); (b) leaves in \emph{tree-like} queries can be anchors or existential (unanchored) variables (here $w$); (c) arbitrary pattern queries can have cycles.
}
\label{fig:query_patterns}
\end{figure}

This observation has spurred the development of numerous machine learning approaches to query answering \citep{DBLP:conf/nips/HamiltonBZJL18,query2box,ArakelyanDMC21,conE,zhu2022neural,ArakelyanMDCA23}, see \citet{Ren-Survey_2023} for a comprehensive recent survey. However, as Ren et al. argue in their survey, most of these approaches involve a query-processing phase that assumes an ordering of the variables, or nodes, of the query, hence defining a traversal of the query that dictates the way it is processed. This poses a challenge for  cyclic queries, where a single traversal does not exists, and thus any strategy assuming the existence of a node ordering cannot faithfully process cyclic queries
\citep{Ren-Survey_2023}. Furthermore, this node traversal is commonly assumed to be \emph{anchored}, in the sense that the traversal starts with particular nodes in the knowledge graph. Figure \ref{fig:query_patterns}(a) shows an example of an anchored tree-like query. Although such queries already capture interesting properties in graphs, they are not capable of checking more complex properties such as the existence of triangles or multiple edges between entities. Examples of more complex query patterns include \emph{tree-like} query patterns in which leaf nodes may be unanchored (a variable), shown in Figure \ref{fig:query_patterns}(b), and cyclic query patterns, as shown in Figure \ref{fig:query_patterns}(c).
\looseness=-1

The development of 
machine learning approaches for more complex query classes remains largely unexplored. In particular, supporting cyclic pattern queries, such as the triangle query, has been identified as an important open challenge by \citet{Ren-Survey_2023}.

Our main contribution  is a \textbf{framework to answer complex graph pattern queries, including cyclic patterns and tree-like queries without anchors}. Our framework consists of two key parts. First, cyclic pattern queries are \emph{unraveled} into complex tree-like queries that capture as much information as possible about the original query. Then, these query unravelings are processed by a neuro-symbolic query processor capable of answering them. More specifically, our \textbf{contributions} are as follows:

\textbf{(1)} We develop \qta, a trainable algorithm that processes cyclic queries via an unraveling strategy. This strategy serves as an approximation scheme with strong \emph{theoretical guarantees}: it is \emph{safe}, meaning no false negative query answers are produced, and it is \emph{optimal} in that we provide the best possible approximation using tree-like queries whenever such an approximation exists.

\textbf{(2)} \qta is \emph{adaptive} in that it is parameterized by the notion of \emph{depth} of tree-like pattern queries. For any depth, an unraveling exists, and higher depth queries potentially offer better approximations. The choice of depth can be tuned based on available resources, queries, and data.

\textbf{(3)} Our neuro-symbolic processor for unravelings can answer any tree-like query \emph{independently of its anchors}. We achieve this by quantifying existential quantification directly in the latent space, which may be of independent interest for developing better processors for tree-like queries. Our algorithm can also process tree-like queries that use negation or union in their edges, even if not anchored, hence offering a strict improvement on the type of queries usually handled by neuro-symbolic techniques. 

\section{Preliminaries}\label{sec:prelims}

\paragraph{Knowledge graphs.}

Knowledge graphs are directed graphs with labeled edges. Formally, a \emph{knowledge graph} is represented as a triple $\G = (\V,\E,\R)$ where $\V$ is a finite set of \emph{entities}, $\R$ is a finite set of \emph{edge types}, and $\E\subseteq \V\times \R\times \V$ is a finite set of \emph{edges}. An edge $(a,R,b)$ is typically denoted by $R(a,b)$. Additionally, $b$ is considered connected to $a$ via the inverse relation $R^{-1}$, denoted as $(b,R^{-1},a)$ or $R^{-1}(b,a)$. The set $\R^{-1}$ represents the inverse relations of $\R$.

\paragraph{Graph pattern queries.}
A \emph{graph pattern query} $q$ on knowledge graphs is specified by a directed labeled graph, where:
(i)~each node is labeled with a unique \emph{constant} (entity from $\V$) or variable ($x,y,z$, etc.); and (ii)~each edge is labeled with one edge type from $\R$. A crucial constraint is that each variable can only be assigned to a single node, while constants can adorn multiple nodes.\footnote{We use a slightly more general definition compared to what is often used in related literature. In our case, multiple nodes in the \emph{query} graph can be denoted with the same constant.}
Furthermore, a single variable must be declared as the \emph{target variable} of $q$, denoted as $q(x)$. Figure~\ref{fig:query_patterns} depicts three graph pattern queries, each with $x$ as target variable.

A graph pattern query $q(x)$ is termed \emph{tree-like} if it forms a tree rooted at node $x$ when ignoring edge direction. Specifically, multiple edges between pairs of nodes are not allowed. A tree-like query $q(x)$ must also satisfy that its non-leaf nodes are only labeled by variables. Moreover, $q$ is \emph{anchored} if all the leaves of this tree are constants; otherwise, it is \emph{unanchored}. The \emph{depth} of a tree-like query refers to the depth of the tree, indicating the length of the longest path from the root to any of its leaves. Finally, $q$ is \emph{cyclic} if it contains a cycle, when ignoring edge direction. The query shown in Figure~\ref{fig:query_patterns}(c) is cyclic, while those shown in Figures~\ref{fig:query_patterns}(a) and \ref{fig:query_patterns}(b) are tree-like and of depth two. In addition, the query in Figure~\ref{fig:query_patterns}(a) is anchored, and the query in Figure~\ref{fig:query_patterns}(b) is unanchored.

It is worth mentioning that graph pattern queries can also be described using first-order logic formulas with atomic predicates, conjunction, and existential quantification (cf. Appendix~\ref{sec:app_prelim}). For illustration, Figure~\ref{fig:query_patterns} also shows the logical formula of each graph pattern query.

\paragraph{Query answering.}
Given a graph pattern query $q(x)$ and a knowledge graph $\G$, an \emph{embedding} of $q$ in $\G$ is a mapping $\mu$ from variables in $q$ to constants in $\V$ such that each edge $e$ in $q$ maps to an edge $\mu(e)$ in $\G$. Here, $\mu(e)$ is derived by replacing variable nodes with constants as determined by $\mu$. The \emph{answers} of $q(x)$ on $\G$ constitute the set $q(\G) := \{\mu(x) \in \V \mid \text{$\mu$ embeds $q$ in $\G$}\}$. These are also referred to as the \emph{easy answers}, as they are derived solely from the information available in $\G$. In contrast,
 neuro-symbolic approaches aim to find \emph{hard answers}, which rely on unknown or missing information in the knowledge graph \citep{Ren-Survey_2023}. 

\section{\unr: Query approximation by unraveling}\label{sec:qta}

In this section, we present our architecture, named 
Unravel (\unr).
Following previous work, our architecture is neuro-symbolic: the task of predicting edges between nodes is carried out with trainable modules, and those modules are linked together with operations that mimic or approximate the symbolic operations mandated by the query. Next, we explain the two main components of \unr:
how we process tree-like queries and how to extend this architecture to handle arbitrary graph patterns by unraveling queries into tree-like queries.

\subsection{Processing tree-like queries}
The basic component of our architecture is a trainable neural model designed to tackle tasks such as link prediction and other general tasks. Our architecture is based on $\gnnqe$ \citep{zhu2022neural,galkin2022inductive}, but it has been extended so that it supports any unanchored tree-like query. It employs a bottom-up approach to process queries, iteratively handling subqueries while maintaining a feature-vector representation that signifies the likelihood of each node in the graph belonging to the answer set of the respective subquery.

Similar to the methodology outlined in \citet{zhu2022neural}, the only trainable components of our architecture are the Neural Bellman-Ford Networks (\nbfnets) \citep{zhu2021bellman}, with each relation and inverse relation in the knowledge graph associated with a dedicated \nbfnet. For a given relation or inverse relation $R \in \R \cup \R^{-1}$, the corresponding \nbfnet, denoted as $\cP_R$, operates as follows: it takes as input a vector $\bx \in [0,1]^{|\V|}$, which reflects the likelihoods of nodes, and returns the updated likelihoods $\cP_{R}(\bx) \in [0,1]^{|\V|}$ after traversing the graph using edges governed by $R$. 
For tree-like queries, we initiate the process from initial vectors in $[0,1]^{|\V|}$ at the leaves. In previous works, only anchored leaf nodes were considered. In that case, if $a$ is the constant in the leaf node, then the indicator vector $\bm{1}_a\in [0,1]^{|\V|}$ is used as initialization. Here,   $\bm{1}_a(a)=1$ and $\bm{1}_a(b)=0$ for all $b\neq a$. Intuitively, \nbfnets are used here to complete $\G$ by adding predicted edges.

Queries whose leaves are (existential) variables are not supported by $\gnnqe$. We handle these queries by encoding them directly in the latent space with the "all-ones" vector $\bm{1}\in  [0,1]^{|\V|}$, as a way to model a uniform prior over all entities.
Once the leaf node vectors are all initialized, we propagate these vectors upwards using the \nbfnets corresponding to the relations. Moreover, if $\bx_1$ and $\bx_2$ are vectors in $[0,1]^{|\V|}$ obtained from processing sub-queries $q_1$ and $q_2$, respectively, then the output vector for the conjunction $q_1 \land q_2$ (merging of the two query graphs) is obtained as the pointwise multiplication $\bx_1 \odot \bx_2$. The final vector and evaluation of the query is the vector returned at the root node labeled with the target variable of the query.

\begin{example}
\label{exa-qta-1}
Consider the tree-like query shown in Figure \ref{fig:query_patterns}(a). Let $\cP_{\texttt{Employee}}$ and $\cP_{\texttt{Friend}}$ be the trained \nbfnets for the relations in our graph. The two leaves in the query are anchored with constants "Tech" and "Bank", and we initialize them with the indicator vectors $\bm{1}_{\text{Tech}}$ and $\bm{1}_{\text{Bank}}$, respectively. We traverse upward by computing $\cP_{\texttt{Employee}}(\bm{1}_{\text{Tech}})$ and $\cP_{\texttt{Employee}}(\bm{1}_{\text{Bank}})$. Intuitively, the $i$th entry of $\cP_{\texttt{Employee}}(\bm{1}_{\text{Bank}})$ represents the likelihood of an edge $(\text{Bank},\texttt{Employee},n_i)$, with $n_i$ being the $i$th node in the graph. The traversal continues by applying $\cP_{\texttt{Friend}}$ to both $\cP_{\texttt{Employee}}(\bm{1}_{\text{Tech}})$ and $\cP_{\texttt{Employee}}(\bm{1}_{\text{Bank}})$. Finally, both paths in the tree are joined when reaching the target variable $x$. This conjunction (join) corresponds to computing
\[    
\bigl(\cP_{\texttt{Friend}}(\cP_{\texttt{Employee}}(\bm{1}_{\text{Tech}}))\bigr)\odot\bigl(\cP_{\texttt{Friend}}(\cP_{\texttt{Employee}}(\bm{1}_{\text{Bank}}))\bigr)
\]
which represents the vector in $[0,1]^{|\V|}$ indicating the likelihood of nodes to be in the query answer.

Next, consider the tree-like query shown in Figure \ref{fig:query_patterns}(b) in which one of the leaf nodes is unanchored. As mentioned, this corresponds to an initialization with $\bm{1}$. In the same way as before, the query is evaluated using
$
\bigl(\cP_{\texttt{Friend}}(\cP_{\texttt{Employee}}(\bm{1}))\bigr)\odot\bigl(\cP_{\texttt{Friend}}(\cP_{\texttt{Employee}}(\bm{1}_{\text{Bank}}))\bigr)$, which returns again a vector in $[0,1]^{|\V|}$ encoding likelihoods of nodes to be in the query answer.
\eox
\end{example}

\begin{figure}
\centering
\includegraphics[height=3.6cm]{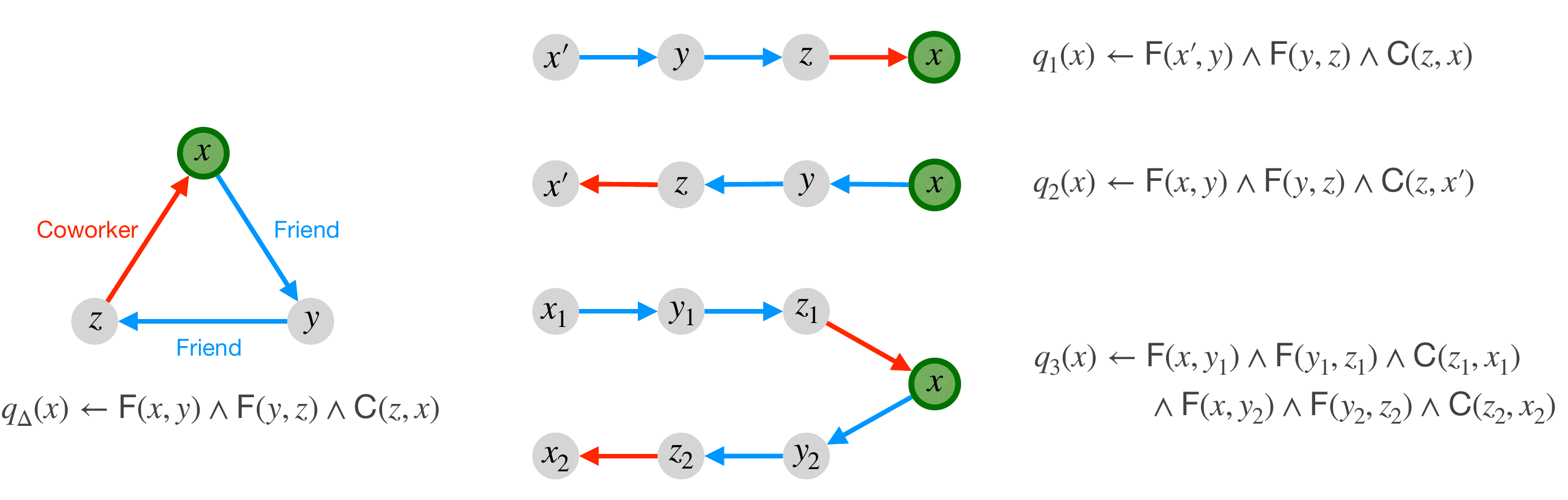}
\caption{The triangle query $q_\Delta$ and tree-like approximations. Best viewed in color. }
\label{fig:approximations}
\vspace{-1ex}
\end{figure}

\subsection{Processing general graph pattern queries} 

To support queries with cycles, we introduce an approach based on computing \emph{traversals} of a query. The traversals we compute are (not necessarily anchored) tree-like queries, and we deal with them using the evaluation strategy outlined in the previous section.
As such, 
\unr is one of the first architectures to provide a neuro-symbolic evaluation method for 
cyclic graph pattern queries.

\begin{example}\label{ex:cyclic_q}
Consider the triangle query $q_\Delta$ shown in Figure \ref{fig:approximations}~(left). We can associate with $q_\Delta$ various tree-like queries, obtained by \emph{traversing} the edges (ignoring direction) in $q_\Delta$, starting from the target variable $x$. For example, the pattern queries $q_1$, $q_2$, and $q_3$ shown in Figure \ref{fig:approximations}~(right) are examples of traversals of $q_\Delta$. Let $a$ be an answer in $q_\Delta(\G)$ for some knowledge graph $\G$. An important property of such traversals is that $a$ will also belong to $q_1(\G)$, $q_2(\G)$, and $q_3(\G)$, simply because there must be a traversal of edges $(a,\texttt{Friend},b),(b,\texttt{Friend},c),(c,\texttt{Coworker},a)$ in $\G$. We say that $q_\Delta$ is \emph{contained} in $q_1,q_2$ and $q_3$. This property holds true for any answer on any knowledge graph. The opposite is not necessarily true, e.g., $a' \in q_1(\G)$ does not necessarily imply that $a' \in q_\Delta(\G)$. \eox
\end{example}
Motivated by the example, we \emph{over-approximate} cyclic pattern queries by their traversals. As the example shows, there may be many such traversals. However, our framework \unr  only uses the \emph{best traversal}, called the \emph{unraveling} of the query.
The construction is parameterized by a \emph{depth} parameter that poses a trade-off: the greater the depth, the closer the unraveling approximates a query, but the costlier it is to compute. 

Algorithm \ref{alg:cap} explains how to compute the unraveling. Starting with the target variable of $q$, we compute its local unraveling by moving through the edges of the query and recursively unraveling along the variables we find. Importantly, further calls to local unraveling also pass on the edge used to arrive at the variable being unraveled, which will not be traversed in that particular call; this avoids traversing the same edge consecutively. The output is always a tree-like query since different recursive calls, when computing the local unraveling of a variable,  involve disjoint fresh variables.

For a given query $q$, let us denote by $\unrav{q}{d}$ its \emph{unraveling of depth $d$}, as computed by Algorithm~\ref{alg:cap}. Our architecture \unr will, on input query $q$ and graph $\G$, consider its tree-like unraveling  $\unrav{q}{d}$ and
use the bottom-up evaluation technique, described earlier,
to approximate $q$.
\begin{algorithm}[t] 
\caption{Unraveling of a general graph pattern query}\label{alg:cap}
\begin{algorithmic}
\Require Depth $d$, query $q(x)$ with edges $\{ R_i(x_i,y_i)
\}_{i=1}^n$ and target variable $x$.
\Ensure Unraveling of $q$ at depth $d$.
\State \textbf{Return} Local unraveling of variable $x$ in query graph $\{R_i(x_i,y_i)\}_{i=1}^n$ at depth $d$, and make $x$ the target variable.
\end{algorithmic}
\end{algorithm}

\begin{algorithm}[t!]
\caption{Local unraveling}\label{alg:cap_local}
\begin{algorithmic} 
\Require Depth $d$, query graph $q=\{R_i(x_i,y_i)\}$, variable $x$ and an optional edge $e$ of $q$.
\Ensure Local unraveling of variable $x$ in $q$ at depth $d$, without using $e$.
\If{$d$ = 0}
    \State return the empty query. 
\EndIf
\State $\In(q,x)\gets \{R(y,x) \mid R(y,x) \text{ is an edge in $q$}, R(y,x) \neq e\}$.
\State $\Out(q,x)\gets \{R(x,y) \mid R(x,y) \text{ is an edge in $q$}, R(x,y) \neq e\}$.
\State $\hat q \gets $ empty query. 
\For{each edge $f$ in $\In(q,x)\cup\Out(q,x)$ }
    \If{$f=R(x,y)$ \textbf{or} $f=R(y,x)$ with $y$ a constant}
        \State add the edge $f$ to $\hat q$.
    \Else 
        \State add to $\hat q$ the edge $f'$, obtained from $f$ by replacing variable $y$ by a fresh variable $y'$. 
        \State $\alpha \gets$ local unraveling of variable $y$ in $q$ at depth $d-1$, without using $f$.
        \State $\alpha' \gets $ replace in $\alpha$, variable $y$ by $y'$, and every other variable by a new fresh variable.
        \State add to $\hat q$  every edge in $\alpha'$.
    \EndIf
\EndFor
\State \textbf{Return} $\hat q$. 
\end{algorithmic}
\end{algorithm}

\begin{example}
Let us compute the unraveling of depth $3$ for a slightly more general triangle query $q_\Delta(x)$ consisting of edges $q=\{R(x,y),S(y,z),T(z,x)\}$. For depth $d=3$, the local unraveling  at $x$ adds edges $R(x,y_2)$ and $T(z_1,x)$ to $\hat q$, with appropriately renamed variables, originating from  $\In(q,x)$ and $\Out(q,x)$. The two recursive calls for the depth $d=2$ local unravelings of $\{S(y,z),T(z,x)\}$ at $y$, and of $\{R(x,y),S(y,z)\}$ at $z$, respectively, add the edges $S(y_2,z_2)$ and $S(y_1,z_1)$ to $\hat q$. In this example, we end up with two additional calls for the depth $d=1$ local unravelings of $\{R(x,y),T(z,x)\}$ at $z$ and at $y$, respectively, adding edges $T(z_2,x_2)$ and $R(x_1,y_1)$ to $\hat q$. The final unraveling is the tree-like query $\hat q(x)$ with edges $\{R(x,y_2),S(y_2,z_2),T(z_2,x_2),T(z_1,x),S(y_1,z_1),R(x_1,y_1)\}$. This unraveling corresponds to $q_3(x)$, shown in Figure \ref{fig:approximations}~(right), when $R$ and $S$ are \texttt{Friend} and $T$ is \texttt{CoWorker}. Hence, for the depth $3$ approximation of $q_\Delta$, \qta returns     
\[
\bigl(\cP_{\texttt{CoWorker}}\bigl(\cP_{\texttt{Friend}}\bigl(\cP_{\texttt{Friend}}(\bm{1})\bigr)\bigr)\odot
\bigl(\cP_{\texttt{Friend}^{-1}}\bigl(\cP_{\texttt{Friend}^{-1}}\bigl(\cP_{\texttt{CoWorker}^{-1}}(\bm{1})\bigr)\bigr)
\]
for the likelihoods of answer nodes. 
\eox
\end{example}

\subsection{Theoretical guarantees}\label{subsec:theory}

In this section we state the most important properties of unravelings, see Appendix~\ref{sec:app_qta_theory} for more details. The first two properties show that our algorithm does produce good approximations of queries. 

\begin{proposition}\label{prop:unravel_prop_safe}
For any query $q$, its unraveling $\unrav{q}{d}$ at depth $d$ satisfies:
\begin{itemize}
    \item (\textbf{Safety}) It is an over-approximation of $q$:  
    For any knowledge graph $\G$, any answer to $q$  on $\G$ is always an answer to $\unrav{q}{d}(\G)$ on $\G$.
    \item (\textbf{Conservativeness}) If $q$ is a tree-like query of depth $d$, then $\unrav{q}{d}$ is equivalent to $q$. That is, they produce the same answers on any knowledge graph.
     \eop
\end{itemize}
\end{proposition}

Our third property states that $\unrav{q}{d}$ is the \emph{best possible tree-like approximation} at depth $d$ of any query $q$ without constants.

 \begin{proposition}\label{prop:unravel_prop_opt}
For any query $q$ without constants, its unraveling $\unrav{q}{d}$ at depth $d$ satisfies:
\begin{itemize}
    \item (\textbf{Optimality}) 
    It is the best possible over-approximation: any other tree-like query $q'$ of depth $d$ that satisfies the safety property must always produce the same or more answers that $\unrav{q}{d}$ (but not less). That is, 
    we have that $\unrav{q}{d}(\G)$ is contained in $q'(\G)$ for any graph $\G$.\eop
\end{itemize}
\end{proposition}
The reason we only show optimality for queries without constants is that the unraveling of queries with constants have uneven depth, hence opening up room for intricate approximations that end up being incomparable to our unraveling. 

Finally, we also have that higher depth unravelings potentially provide better approximations. The choice of depth can be tuned depending on available resources, queries and data at hand.
\begin{proposition}\label{prop:unr_depth}
For any query $q$ and $d>0$, the set of answers $\unrav{q}{d}(\G)$ is always contained in the set of  answers  $\unrav{q}{d-1}(\G)$ on any knowledge graph $\G$.\eop
\end{proposition}

\section{Experiments}
\label{sec-experiments}

Here, we empirically investigate the performance of \unr for answering complex graph pattern queries. Specifically, we aim to address the following questions:

\textbf{Q1:} How does \unr perform on cyclic queries?\\
\textbf{Q2:} How does the depth of the unraveling affect the quality of the \unr approximation?\\
\textbf{Q3:} How does \unr perform on tree-like queries? Is it able to answer unanchored queries whilst remaining competitive on anchored queries against other neuro-symbolic approaches?

To answer \textbf{Q1}, we compare against $\mpqe$ \citep{daza2020mpqe}, a neural query answering approach capable of dealing with cyclic patterns. We also build a new baseline, \baseline,  that does not rely on unravelings and is trained on link prediction rather than querying. For \textbf{Q2}, note that unlike theoretical guarantees, longer queries can indeed perform worse because our trainable methods may not be geared at computing them. Hence, we zoom in on results for different depths of unraveling. For \textbf{Q3}, we train $\unr$ on both anchored and unanchored queries, and compare to $\gnnqe$ \citep{zhu2022neural}, since it is the method closest to our tree-like query processor. This allows us to compare how adding unanchored queries in training helps or hinders the ability to answer anchored tree-like queries ($\gnnqe$ is only trained on anchored queries because it cannot handle unanchored ones). 

The source code for \unr and other information are available at 
\url{https://anonymous.4open.science/r/UnRavL-A6EF/}. 

\subsection{Experimental setup}
We perform our evaluation on the commonly used knowledge graphs FB15k-237 \citep{toutanova2015observed}, FB15k \citep{bordes2013translating} and NELL995 \citep{Nell995} with their official dataset splits. 
That is, it is trained using the queries generated by BetaE \citep{ren2020beta}, consisting of 10 anchored tree-like query types, including queries featuring union and negation ($1$p/$2$p/$3$p/$2$i/$3$i/$2$in/$3$in/inp/pni/pin), see Figure~\ref{fig:Renbetaqueries} in the Appendix for a graphic depiction of queries.
For our \qta method, we additionally provide a new set of training, validation and test queries without anchors with their corresponding answers and test unravelings of cyclic queries for FB15k-237, FB15k and NELL995.
We indicate unanchored query types by preceding it by an $\exists$ (for existentially quantified). For example, $\exists 2$p
corresponds to queries $q(x)$ with edges of the form $R(z,y),S(y,x)$ with $x,y,z$ all variables. Furthermore, alongside the unanchored \emph{triangle} query ($\exists3$c) and \emph{four cycle} ($\exists4$c) we also consider the \emph{lollipop query} ($\exists1$p$2$c) consisting of edges $R(z,y),S(y,x),T(y,x)$ and target variable $x$. See Figure~\ref{fig:cyclic_queries} in the Appendix for a depiction of cyclic queries. Following the practice used for anchored queries, we use  ($\exists1$p/$\exists2$p/$\exists3$p/$\exists2$i/$\exists3$i/$\exists2$in/$\exists3$in/$\exists$inp/$\exists$pni/$\exists$pin) for training, and left $\exists$ip, $\exists$pi, $\exists2$u, $\exists$up and cyclic queries for testing only. 

We evaluate  how well the predicted answers correspond to true answers.
Following previous work \citep{zhu2022neural}, we use mean reciprocal rank (mrr) and hits@k for this purpose.
Moreover, to understand the intricacies of our model, we also incorporate classification metrics such as precision and recall using different classification thresholds. 

\subsection{Baselines and other approaches}

Our baseline \baseline is based on a non-trainable neuro-symbolic algorithm from the  
\emph{Probabilistic Query Evaluation} literature. \baseline starts by building a probabilistic database by assigning a likelihood score to each potential relation using an \nbfnets link predictor \citep{zhu2021bellman}.  
Then, treating these scores as probabilities, \baseline evaluates the query directly using a probabilistic solver. 
Storing the probabilistic graph requires $\mathcal{O}(\vert\mathcal{V}\vert^2)$ space and it was feasible only for the FB15k-237 dataset, which is also why this approach is often disregarded as a possible baseline \citep{DBLP:conf/nips/HamiltonBZJL18,ArakelyanDMC21}. As for the probabilistic solver, 
given the $\#\text{P}$-hardness of computing the \emph{possible world semantics} of graph pattern queries \citep{DBLP:conf/pods/DalviS07a}, \baseline utilizes a polynomial time evaluation method based on \emph{dissociations} of the original queries \citep{diss3}. This approach is known to provide good upper approximations of the actual probabilities of query answers. 
Further details on \baseline can be found in Appendix \ref{sec:app_base_theory_exp}. 

As a second baseline, we consider Message Passing Query Embedding (\mpqe)~\citep{daza2020mpqe}. \mpqe is a neural method that learns embeddings of entities in the knowledge graph, together with an \emph{encoder} of queries--a graph neural network that maps a query graph to a vector representation. For a given query, \mpqe returns a ranked list of entities by computing the cosine similarity between the query embedding computed by the encoder, and the embeddings of all entities in the graph. Although not studied in the original publication, the query encoder in \mpqe supports arbitrary query patterns, including cyclic queries and unanchored ones. We train \mpqe with anchored and unanchored tree-like queries, and we use the performance on the validation set for hyperparameter tuning. More details about our experimental setup with \mpqe can be found in Appendix~\ref{sec:app-mpqe-details}.

\subsection{Implementation}
As explained, for \baseline we use \nbfnets \citep{zhu2021bellman} to \emph{complete} the graph and materialize such results and compute scores of the dissociated queries. 
The implementation of $\qta$ is built on top of \gnnqe \citep{zhu2022neural}. We adapted \gnnqe to support unanchored tree-like queries, as explained earlier in the paper. Further, we train using both anchored and unanchored tree like queries. 
For \mpqe we rely on the implementation by~\citet{alivanistos2021query}, which makes use of a \textsf{CompGCN}~\citep{VashishthSNT20} for the query encoder instead of the original \textsf{R-GCN}.
In line with previous works \citep{query2box,ren2020beta,conE,zhu2022neural}, our model (that is, the underlying \nbfnets) is trained to minimize the binary cross entropy loss. Full details of the architecture and training of both \nbfnet and \gnnqe can be found in Appendix \ref{sec:app-training-implementation}. 
\subsection{Results}

\begin{table}[t]
\caption{Results of $\qta$, $\baseline$ and $\mpqe$ for cyclic queries: lollipop query ($\exists$1p2c), triangle query ($\exists$3c) and square query ($\exists$4c). Results for $\qta$ correspond to the best unraveling depth, while results for $\baseline$ corresponds to the metrics when considering the smallest dissociation score for each entity.}
    \label{tab:tree-like-base-qta-mpqe}
    \centering
    \resizebox{0.8\textwidth}{!}{
    \begin{tabular}{ccccccccccc}
    \toprule
    & & \mc{3}{c}{FB15k237} & \mc{3}{c}{FB15k} & \mc{3}{c}{NELL995} \\
    \cmidrule(lr){3-5}   \cmidrule(lr){6-8} \cmidrule(lr){9-11}
    \bf Metric                     & \bf Model            & $\pmb\exists$\textbf{1p2c} & $\pmb\exists$\textbf{3c} & $\pmb\exists$\textbf{4c}  & $\pmb\exists$\textbf{1p2c} & $\pmb\exists$\textbf{3c} & $\pmb\exists$\textbf{4c}  & $\pmb\exists$\textbf{1p2c} & $\pmb\exists$\textbf{3c} & $\pmb\exists$\textbf{4c}     \\
    \midrule
\multirow{3}{*}{hits@1}
& $\qta$   &  \bf 0.027  & \bf 0.073   & \bf 0.036   & 0.057  & \bf 0.092   & \bf 0.057 & 0.065  & \bf 0.114   & 0.067    \\
& $\baseline$ &  0.011  & 0.027   & 0.009  & --- & --- & ---& --- & --- & ---        \\ 
& $\mpqe$ &   0.023 & 0.031 & 0.005 & \bf 0.064 & 0.063 & 0.000 & \bf 0.074 & 0.060 & \bf 0.098 \\ \midrule
\multirow{3}{*}{mrr}
& $\qta$   &  0.067  & \bf 0.148   & \bf 0.088 & 0.098  & \bf 0.297 & \bf 0.159  & 0.146  & \bf 0.211   & 0.122    \\
& $\baseline$ &  0.057  & 0.077   & 0.031 & --- & --- & ---& --- & --- & ---         \\ 
& $\mpqe$ & \bf 0.069 & 0.072 & 0.025 & \bf 0.120 & 0.116 & 0.000 & \bf 0.150 & 0.142 & \bf 0.180 \\
\bottomrule
\end{tabular}
}
\end{table}

To present our results we address each of \textbf{Q1}, \textbf{Q2} and \textbf{Q3} separately. 

\paragraph{Q1 - Cyclic queries.} 
Table \ref{tab:tree-like-base-qta-mpqe} displays results for cyclic queries on FB15k-237, FB15k, and NELL995. 
The results for $\baseline$ correspond to the dissociation and results for $\qta$ correspond to the best unraveling depth.
In FB15k-237, we see that \baseline is the worst performer, highlighting the advantage of training with different types of queries, and not just links. 
Moreover, \unr outperforms \mpqe on the longer queries $\exists$3c and $\exists$4c, with better performance on 8 out of 9 query / dataset pairs, and \mpqe is the best performer on the shorter $\exists$1p2c query. 
Interestingly, \unr shows better performance in all queries over all datasets when amplifying our metric to hits@10 (see Appendix \ref{sec:app-cyclic-fb15k-nell}).

\begin{table}[ht]
\centering
\caption{Results of $\qta$ for the triangle  query's unravelings on FB15k-237 dataset, depths 2 to 10}
\label{table:depths-triangle}
\vspace{0.2cm}
\resizebox{0.85\textwidth}{!}{%
\begin{tabular}{llllllllllllll}
\label{tab:triangle-unravels}
\textbf{Depth} & \textbf{} & \textbf{2} & \textbf{3} & \textbf{4} & \textbf{5} & \textbf{6} & \textbf{7} & \textbf{8} & \textbf{9} & \textbf{10}\\ \hline
mrr                    &           & 0.137 &    0.147       &    0.143        &    0.140        &     0.143       &      0.141      &   0.143  & \bf 0.148 & 0.147                  \\   
hits@1                    &  &    0.066       &    0.068       &    0.071        &    0.068        &     0.072      &      0.070      &   0.068 & 0.073 & \bf 0.075          \\
hits@10                    &  &    0.285      &    \bf 0.301       &    0.290        &    0.290        &     0.287      &      0.283      &   0.292 & 0.292 &  0.293          \\
\end{tabular}%
}
\end{table}
\renewcommand{\arraystretch}{1.1}
\begin{table}[ht]
\centering
\caption{Results of $\qta$ for the square query's unravellings on FB15k-237 dataset, depths 3 to 12}
\label{table:depths-square}
\vspace{0.2cm}
\resizebox{0.9\textwidth}{!}{%
\begin{tabular}{lllllllllllllll}
\label{tab:square-unravels}
\textbf{Depth} & \textbf{} & \textbf{3} & \textbf{4} & \textbf{5} & \textbf{6} & \textbf{7} & \textbf{8} & \textbf{9} & \textbf{10} & \textbf{11} & \textbf{12} \\ \hline
mrr                    &           & 0.074&    \bf 0.088      &    0.085        &    0.087        &     0.087       &      0.087      &   0.087  & 0.087 & \bf 0.088 & 0.086                \\   
hits@1                    &           & 0.023&    \bf 0.035      &    0.028        &    0.033        &     0.033       &      0.034      &   0.034  & 0.032 & \bf 0.036 & 0.030                  \\ 
hits@10                    &           & 0.166&    0.187      &    0.193        &    0.197        &     \bf 0.201       &      0.178      &   0.181  & 0.188 &  0.182 & 0.192                  \\ 
\end{tabular}%
}
\end{table}
\renewcommand{\arraystretch}{1.1}
\begin{table}[b]
\caption{Results of $\qta$, $\baseline$ and $\mpqe$ for unanchored tree-like queries on FB15k-237, FB15k, and NELL995.} 
\label{tab:tree-like-base-qta}
\vspace{0.2cm}
\resizebox{\textwidth}{!}{%
\begin{tabular}{llllllllllllllll}
\bf Metric                     & \bf Model            & $\pmb\exists$\textbf{1p} & $\pmb\exists$\textbf{2p} & $
\pmb\exists$\textbf{3p} & $\pmb\exists$\textbf{2i} & $\pmb\exists$\textbf{3i} & $\pmb\exists$\textbf{ip} & $\pmb\exists$\textbf{pi} & $\pmb\exists$\textbf{2in} &$\pmb\exists$\textbf{3in} & $\pmb\exists$\textbf{inp} & $\pmb\exists$\textbf{pin} & $\pmb\exists$\textbf{pni} & $\pmb\exists$\textbf{2u} & $\pmb\exists$\textbf{up} \\ \hline
\multicolumn{16}{|c|}{\textbf{General tree-like queries FB15k-237}}  \\ \hline
\multirow{3}{*}{hits@1}    & $\qta$          &  \bf 0.028  & \bf 0.046   & \bf 0.057   & \bf 0.123   & \bf 0.203   &  \bf 0.030  & \bf 0.090   &  \bf 0.011   &  \bf 0.055   &   \bf 0.028  &  \bf 0.019   & \bf 0.005    & \bf 0.030   &  \bf  0.023  \\

& $\baseline$ &  0.021  & 0.001   & 0.011   & 0.017   & 0.071   &  0.027  & 0.051   &  0.007   &  0.036   &   0.022  &  0.010   &  0.003   &  0.021  &  0.011  \\ 

& $\mpqe$ &  0.012 & 0.041 & 0.035 & 0.021 & 0.027 & \bf 0.030 & 0.028   & - & - & - & - & - & - & - \\ \hline
\multirow{3}{*}{mrr}       & $\qta$           &  \bf 0.055  & \bf 0.088   &  \bf 0.079  &  \bf 0.212  &  \bf 0.292  &  0.076  &  \bf 0.151  &   0.029  &  \bf 0.107   &  \bf 0.061   &   \bf 0.045  &  \bf 0.027   & \bf 0.065   &   \bf 0.053 \\
& $\baseline$ &  0.031  &  0.025  &  0.051  &  0.034  &  0.111  & \bf 0.107   &  0.064  &  \bf 0.051   &  0.057   &   \bf 0.061  &  0.035   & \bf 0.027   &  0.032  &  0.029  \\ 
& $\mpqe$ & 0.029 & 0.069 & 0.061 & 0.041 & 0.049 & 0.052 & 0.046 & - & - & - & - & - & - & - \\ \hline
\multicolumn{16}{|c|}{\textbf{FB15k}}                                                                                    \\ \hline
\multirow{2}{*}{hits@1} & $\qta$ &  \bf 0.831  &  \bf 0.643  &  \bf 0.533  &  \bf 0.759  & \bf 0.791  & \bf 0.566  &  \bf 0.668  &   0.308  &  0.297   &  0.315   &  0.189   &   0.219   &   0.704  &   0.540  \\
& \mpqe & 0.258 & 0.061 & 0.060 & 0.094 & 0.121 & 0.081 & 0.096 & - & - & - & - & - & - & - \\ \hline
\multirow{2}{*}{mrr} & $\qta$ &   \bf  0.855        &   \bf  0.688        &      \bf 0.587       &       \bf0.801      &    \bf   0.833      &    \bf    0.620     &     \bf 0.720        &      0.430        &     0.418        &      0.403        & 0.302        &     0.340         &   0.747         &  0.600   \\ 
& \mpqe & 0.362 & 0.107 & 0.100 & 0.153 & 0.198 & 0.138 & 0.145 & - & - & - & - & - & - & -  \\ \hline
\multicolumn{16}{|c|}{\textbf{NELL995}}                 \\ \hline
\multirow{2}{*}{hits@1}  & $\qta$ & \bf 0.831  &  \bf 0.643  &  \bf 0.533  &  \bf 0.759  & \bf 0.791  & \bf 0.566  & \bf 0.668  &   0.308  &  0.297   &  0.315   &   0.189   &  0.219   &  0.704  &  0.540   \\ 
& \mpqe & 0.191 & 0.032 & 0.046 & 0.066 & 0.114 & 0.031 & 0.056 & - & - & - & - & - & - & - \\ \hline
\multirow{2}{*}{mrr} & $\qta$    & \bf 0.479       & \bf 0.160       & \bf 0.119       & \bf 0.378       & \bf 0.484       & \bf 0.140       & \bf 0.269       & 0.086        & 0.128        & 0.111        & 0.055        & 0.053       & 0.141       &  0.107 \\ 
& \mpqe & 0.270 & 0.061 & 0.075 & 0.132 & 0.198 & 0.060 & 0.097 & - & - & - & - & - & - & - \\ \hline
\end{tabular}%
}
\end{table}

\paragraph{Q2 - Depth.} Tables \ref{table:depths-triangle} and \ref{table:depths-square} show how our approximation scheme performs on the triangle and  square query respectively, when considering different depths for the unravelings. We note that both the mean reciprocal rank and hits@k metric tend to remain relatively stable. Theoretically, the deeper the unravelings, the more precise the results should be (cfr. Proposition~\ref{prop:unr_depth}). However, this is not necessarily captured by either mrr or hits@k, as there some  practical intricancies in play: the training sets for complex query answering models only consider queries that involve up to 3 successive projections (3p), while our unravelings can be arbitrarily deep, which may lead to successive stacking of errors. In Appendix \ref{appendix:more-results} we show that Precision and Recall metrics are more affected by the depth of unravelings, although these metric themselves depend on a particular classification threshold.

\paragraph{Q3 - Tree-like queries.} Table \ref{tab:tree-like-base-qta} shows the performance of \qta, \baseline, and \mpqe (which does not support negations or disjunctions) for general (unanchored) tree-like queries where \qta outperforms both of the other approaches. Importantly, we note that this performance can be achieved without a noticeable drop in performance for \emph{anchored} tree-like queries. For more details, see Appendix \ref{appendix:anchor+unanchored}, where we compare with
\gnnqe. We also report our comparison with \qta trained using anchored queries only.

\section{Related work}\label{sec:relwork}

\paragraph{Neural and neuro-symbolic query answering.}
The machine learning community has produced a wide body of literature investigating how to answer complex queries over incomplete knowledge graphs. These works build on and extend successful methods for learning embeddings of entities and relations in knowledge graphs \citep{bordes2013translating,YangYHGD14a,TrW2016,SunDNT19,SchlichtkrullKB18,VashishthSNT20,TeruDH20}. We can identify two different approaches to complex query answering. Firstly, \emph{neural} approaches \citep{DBLP:conf/nips/HamiltonBZJL18,daza2020mpqe,kotnis2021answering,liu2022mask,ren2020beta,query2box,pflueger2022gnnq} answer queries by designing functions (parameterized by neural networks) for mapping queries and entities in an embedding space where similarity indicates the likelihood of an answer. While competitive, the neural networks in these works act as black boxes that turn a query into a vector, thus resulting in a less interpretable system. Secondly, there are so-called \emph{neuro-symbolic} approaches, which combine neural approaches to compute missing links between entities, and symbolic approaches to extract answers from the completed data~\citep{ArakelyanDMC21, bai2022answering,luo2023nqe,chen2022fuzzy,yin2023rethinking,zhu2022neural,ArakelyanMDCA23}. While logical operators are still processed in the latent space, this process enforces a strong bias when computing queries, as the learning process must take into account the symbolic operations occurring downstream. 
With the exception of \mpqe, both neural and neuro-symbolic methods for query answering in the literature are limited to anchored tree-like queries, which is needed to form a computation graph in an embedding space, and thus they cannot be directly applied to cyclic or unanchored queries. While we consider \mpqe in our experiments, we note that it does not support queries with negations or disjunctions.

\paragraph{Approximation of conjunctive queries.}
The notion of a tree-like approximation of a pattern query, as explored in this paper, was originally introduced by the database theory community. Two types of approximations were proposed: {\em underapproximations}, which yield sound but not necessarily complete answers \citep{DBLP:journals/siamcomp/BarceloL014}, and {\em overapproximations}, which yield complete but not necessarily sound answers \citep{brz20}. 
However, these work tend to concentrate on a slightly different notion of tree-like, namely, \emph{treewidth-1} queries. More importantly, these papers are purely theoretical: there are no previous results about the application of these ideas in practice, let alone their use for neuro-symbolic query answering. 

\paragraph{The notion of unraveling.} 
The idea of unraveling the nodes of a graph to form a tree, also known as tree unfolding has been used before in several domains, from temporal logic \citep{schnoebelen2002complexity} to database theory \citep{buneman2000unql} to formal language theory \citep{vardi1994nontraditional}, and even in machine learning, to formally study methods for graph embeddings \citep{schulz2022generalized}. However, up to our best knowledge, there is no previous work on the practical application of these techniques in the context of traditional query answering or in neural query answering. Hence, our paper also contributes by showing how query approximations obtained by unravelings can be used in practice to provide state of the art querying algorithms.

\section{Conclusions}
\unr improves capabilities of neural query processors with an approximation scheme for cyclic queries, and a latent encoding of existential quantification for queries without anchors. Results show that \unr outperforms other neural methods capable of answering cyclic patterns, as well as a baseline inspired by the area of probabilistic query evaluation. Interestingly, \unr achieves all of this while also remain being competitive for standard anchored, tree-like queries.

Despite our advances, we are still far from obtaining a neural query processing engine capable of dealing with any graph database query \citep{Ren-Survey_2023}. For future work, we plan to continue designing good overapproximation schemes for more complex queries. 
Another line of research is to develop neuro-symbolic methods capable of answering cyclic queries that return exact answers when evaluated on complete data. This process can be computationally demanding, but over the last decade, {\em worst-case optimal} algorithms have been developed for retrieving such answers in a symbolic manner \citep{DBLP:journals/sigmod/NgoRR13}. We plan to investigate how such algorithms can be integrated into the neuro-symbolic framework studied in this paper to provide high-quality answers in a context where data is considered incomplete. 

\section*{Acknowledgments}
Tamara Cucumides is supported by FWO project G019222N. Daniel Daza and Michael Cochez are in part funded by the Elsevier Discovery Lab and Michael Cochez is partially funded by the Graph-Massivizer project, funded by the Horizon Europe programme of the European Union (grant 101093202). Pablo Barceló and Miguel Romero are funded by the National Center for Artificial
Intelligence CENIA FB210017, BasalANID.

\bibliographystyle{plainnat}
\bibliography{main}

\appendix
\begin{figure}
\includegraphics[width=\textwidth]{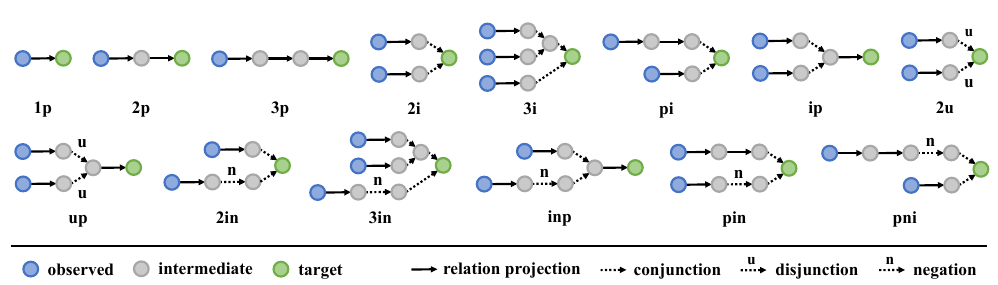}    \caption{Graph pattern queries introduced by \citet{ren2020beta} and used in our experiments. (The figure is taken from \citet{zhu2022neural}.)}
\label{fig:Renbetaqueries}
\end{figure}

\section{Connection to logic}\label{sec:app_prelim}

We have described queries in terms of their graph representation. Here, we make the connection to logic more precise.

Let $\var$ be a countably infinite set of \emph{variables} and let $\con$ be a countably infinite set of \emph{constants}. Consider the following class of first-order logic formulas (called \emph{(unary) conjunctive queries}) over the set $\R$ of edge types and constants in $\con$, of the form (we use rule-based notation)
\[
q(x) \gets R_1(y_1, z_1) \land \cdots \land R_m(y_m, z_m),
\]
where $x$ is the \emph{target} variable, and each $R_i(y_i, z_i)$ is an \emph{atom} with $R_i \in \R$ and $\{y_i, z_i\} \subseteq \con \cup \var$ ($y_i, z_i$ are either variables or constants). The variable set $\var(q)$ of $q$ is the set of variables appearing in the atoms of $q$, that is, the variables appearing in $\{y_1, z_1, \dots, y_m, z_m\}$. Similarly, we denote by $\con(q)$ the constants appearing in the atoms of $q$. As usual, we assume $x \in \var(q)$. The variables in $\var(q) \setminus \{x\}$ are the \emph{existentially quantified} variables of $q$. The semantics of these queries is defined using the standard semantics of first-order logic.

As an example, Figure~\ref{fig:query_patterns}(c) shows the logical formula $q(x) \gets \text{Friend}(x, y) \land \text{Friend}(y, z) \land \text{Coworker}(z, x)$ looking for all persons $x$ that have a friend $y$ and a coworker $z$ who are friends with each other. Here, $\var(q) = \{x, y, z\}$, $x$ is the target variable, and $y$ and $z$ are both existentially quantified.

We turn logical formulas into a graph representation as follows. We take $\var(q) \cup \conocc(q)$ as nodes, and an edge from node $u$ to node $v$ for every atom $R(u, v)$ in $q$. Here, $\conocc(q)$ is the set of occurrences of constants in $q$, i.e., if the number of occurrences in different atoms of $q$ of a constant $a \in \con(q)$ is $k$, then there are $k$ duplicates of $a$ in $\conocc(q)$. The target variable is, of course, taken as the same.

Conversely, it is readily verified that the graph representation of queries corresponds to formulas as described above. Indeed, simply take a conjunction over all edges and existentially quantify all variables except for the target variable.

With this connection in place, we can say that a formula has a certain graph property, e.g., being tree-like, meaning that its graph representation has that property.

\begin{figure}
\centering
\includegraphics[height=3cm]{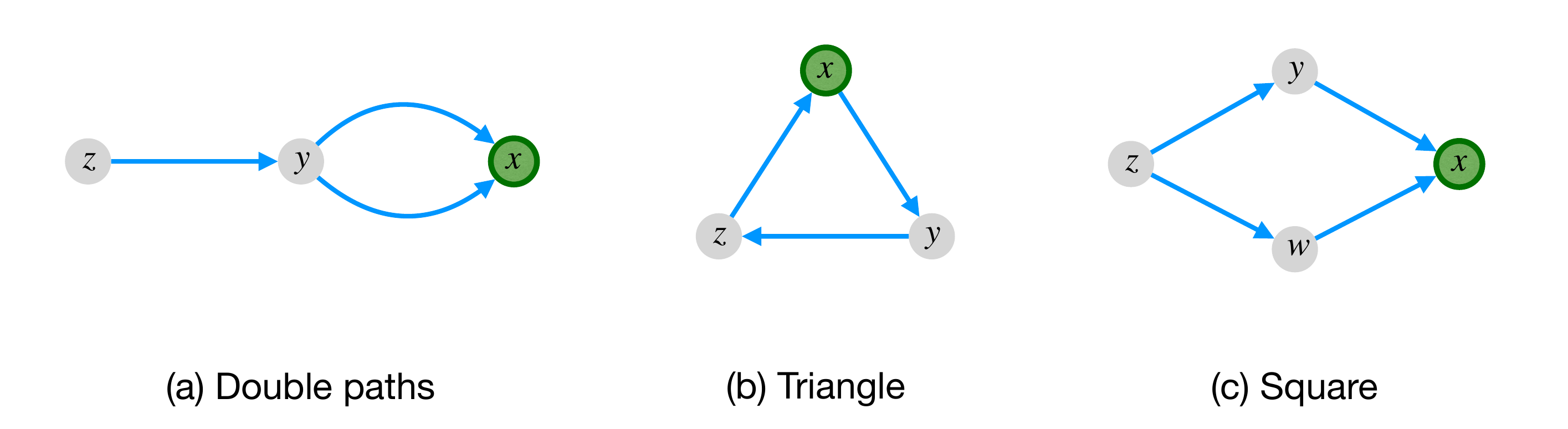}
\caption{Examples of cyclic queries.}\label{fig:cyclic_queries}
\end{figure}
\begin{example}
The cyclic queries shown Figures~\ref{fig:cyclic_queries} can be easily seen to correspond to the logical formulas
$q_{\mathsf{lp}}(x)\gets R(y,x)\land S(y,x)\land T(z,y)$,
$q_{\Delta}(x)\gets R(x,y)\land S(y,z)\land T(z,x)$,
and 
$q(x)\gets R(y,x)\land S(y,x)\land T(z,y)$ and 
$q_{{\resizebox{1ex}{!}{$\square$}}}(x)\gets R(y,x)\land S(w,x)\land T(z,y)\land U(z,w)$, respectively. \eox
\end{example}

\section{Details of Propositions~\ref{prop:unravel_prop_safe},~\ref{prop:unravel_prop_opt} and~\ref{prop:unr_depth}}
\label{sec:app_qta_theory}

We here provide an alternative description for computing the unraveling of a query, which can easily be shown to be equivalent (by induction on $d$) to the unraveling computed by Algorithm~\ref{alg:cap}.

\paragraph{Unraveling in terms of valid paths.}

Let $q(x)$ be a graph pattern query with target variable $x$.  A \emph{valid path} of $q(x)$ is a sequence $u_0, e_1, u_1, \dots, e_k, u_k$, for $k \geq 0$, such that:

\begin{itemize}\setlength{\itemsep}{0pt}%
    \setlength{\topsep}{0pt} 
    \setlength{\partopsep}{0pt}
    \setlength{\parsep}{0pt}
    \setlength{\parskip}{0pt}
    \item For all $1\leq i\leq k$, we have that $u_i$ is a node in the graph pattern. Moreover, $u_0$ is labeled by $x$, $u_i$ is labeled by a variable, for all $1\leq i< k$, and $u_k$ is either labeled by a variable or a constant.
    \item For all $1\leq i\leq k$, we have that $e_i$ is an edge in $q$ with endpoints $u_{i-1}$ and $u_i$ (that is, $e_i$ is either of the form $R(u_{i-1},u_i)$ or $R(u_{i},u_{i-1})$).
    \item For all $1 \leq i < k$, we have that $e_i \neq e_{i+1}$.
\end{itemize}

Intuitively, a valid path is a way of traversing $q$ starting from the target variable $x$ and sequentially moving through the edges of $q$, ignoring direction. We can visit the same variable or edge several times. The only restriction is that an edge cannot be visited multiple times \emph{consecutively} in the sequence. Hence, once an edge is traversed, we cannot go back via the same edge immediately.

The \emph{length} of a valid path is the number of edges $k$. Note that the valid path of length $0$ is well-defined and corresponds to the sequence $x$. A valid path is \emph{unanchored} if it ends at a variable of $q$; otherwise, it is \emph{anchored}. For a valid path $P$, we denote by $\pathend(P)$ the node at the end of path $P$ (which is labeled either by a variable or constant).

\begin{example}
Consider the triangle query $q_\Delta(x)$, shown in Figure~\ref{fig:approximations} (left), consisting of edges $\{\texttt{Friend}(x, y), \texttt{Friend}(y, z), \texttt{Coworker}(z, x)\}$. An example of an unanchored valid path is $x, \texttt{Friend}, y, \texttt{Friend}, z, \texttt{Coworker}, x$, which corresponds to a clockwise traversal of length $3$ starting at $x$. The anticlockwise traversal of length $3$ is given by the valid path $x, \texttt{Coworker}, z, \texttt{Friend}, y, \texttt{Friend}, x$.\eox
\end{example}

Let $q$ be a graph pattern query and $d \geq 1$. We need variables corresponding to each unanchored valid path $P$ of $q$ of length $\leq d$. We denote the variable corresponding to such a path $P$ by a fresh variable $z_P$, although encoding of paths by any fresh variables suffices, as shown next.

\begin{example}
We continue with the triangle query $q_\Delta(x)$. Let us introduce some variables for valid paths:

\[
\begin{array}{c|c}
\text{variable} & \text{path}\\\hline
x & x\\
z_1 & x,\texttt{Coworker},z\\
y_1 & x,\texttt{Coworker},z,\texttt{Friend},y\\
x_1 & x, \texttt{Coworker}, z, \texttt{Friend}, y, \texttt{Friend}, x\\
y_2 & x,\texttt{Friend},y\\
z_2 & x,\texttt{Friend},y,\texttt{Friend},z\\
x_2 & x, \texttt{Friend}, y, \texttt{Friend}, z, \texttt{Coworker}, x\\
\end{array}
\]
We thus have introduced variables for each unanchored valid path 
of length at most three in $q$, starting from $x$.
\eox
\end{example}

The \emph{unraveling} of $q(x)$ of depth $d \geq 1$, as computed by Algorithm~\ref{alg:cap}, is easily shown to be equivalent to the following graph pattern query: 

\begin{itemize}\setlength{\itemsep}{0pt}%
    \setlength{\topsep}{0pt} 
    \setlength{\partopsep}{0pt}
    \setlength{\parsep}{0pt}
    \setlength{\parskip}{0pt}
    \item The nodes of the graph pattern are the valid paths of length $\leq d$. 
    \item Unanchored valid paths $P$ are labeled with the fresh variable $z_P$. Anchored valid paths $P$ are labeled with the constant labeling $\pathend(P)$. 
    \item The target variable is $x:=z_x$.
    \item For valid paths $P$ and $P' = P, e', \pathend(P')$ of $q$ of lengths $\leq d$, if $e'$ is an edge labeled $R$ from $\pathend(P)$ to $\pathend(P')$, then we add an edge with label $R$ from $P$ to $P'$. Otherwise, we add an edge with label $R$ from $P'$ to $P$.
\end{itemize}

We recall that we denote the
corresponding graph query by $\unrav{q}{d}$. 

\begin{example}
It is now easily verified that the graph pattern query $q_3(x)$ shown in Figure~\ref{fig:approximations}(c) is the depth three unraveling of $q(x)$, using the variables for paths introduced in the previous example. Indeed, it is the pattern query consisting of the following edges: $\{\texttt{Friend}(x, y_2), \texttt{Friend}(y_2, z_2), \texttt{Coworker}(z_2, x_2), \texttt{Coworker}(z_1, x), \texttt{Friend}(y_1, z_1), \allowbreak \texttt{Friend}(x_1, y_1)\}$.
\eox
\end{example}

The idea is that the unraveling of depth $d$ of $q(x)$ is obtained by traversing $q$ in a tree-like fashion, starting from the target variable $x$ and moving from one variable to all of its neighbors through the edges of $q$. Each time, we add fresh variables to the unraveling, resulting in a tree-like query. The tree traversal has depth $d$ and is always restricted to valid paths (no immediate returns to the same atom). The leaves of the unraveling could be anchors or variables. As mentioned, the traversals and addition of edges by Algorithm~\ref{alg:cap} precisely follow the description above.

\paragraph{Containment.}
Before proving properties of the unraveling of a query we recall how containment of two queries can be defined in terms of homomorphisms. We recall that a graph pattern query $q$ is \emph{contained} in another graph pattern query  $q'$, denoted by $q\subseteq q'$, if $q(\G)\subseteq q'(\G)$, for all knowledge graphs $\G$. That is, the answer of $q$ is always contained in the answer of $q'$, independently of the underlying knowledge graph. While this notion reasons over all knowledge graphs, it admits a simple syntactic characterization based on homomorphisms. 

A \emph{homomorphism} from query $q(x)$ to query $q'(x)$ is a mapping $h$ from the nodes of $q$ to the nodes of $q'$ such that:
\begin{itemize}
\item $h(x) = x$;
\item if $u$ is a node in $q$ labeled by a constant $a$, then $h(u)$ is also labeled by $a$ in $q'$; and
\item For all edges $R(u,v)$ of $q$, we have that $R(h(u),h(v))$ is an edge of $q'$.
\end{itemize}

Intuitively, a homomorphism is a way of replacing the variables of $q$ by variables of $q'$ such that each edge of $q$ becomes an edge of $q'$. The target variable of $q$ must be mapped to the target variable of $q'$. The following is a well-known characterization of containment of graph pattern queries.

\begin{theorem}[\cite{cm77}]
\label{thm:cm77}
A graph pattern query $q$ is contained in a graph pattern query $q'$ if and only if there is a homomorphism from $q'$ to $q$.\eop
\end{theorem}

\paragraph{Proof of Proposition~\ref{prop:unravel_prop_safe}}
We next provide the proof of Proposition~\ref{prop:unravel_prop_safe}.

\noindent
\underline{$\rhd$ \emph{Safety}}

We start by showing that 
$\unrav{q}{d}$ over-approximates $q$, that is, $q$ is contained in $\unrav{q}{d}$. By Theorem~\ref{thm:cm77} it suffices to show that there exists a homomorphism $h$ from 
$\unrav{q}{d}$ to $q$.

Consider the mapping $h$ from the nodes of $\unrav{q}{d}$ to the nodes of $q$ that maps each valid path $P$ in $\unrav{q}{d}$ to its final element $\pathend(P)$. We claim that $h$ is a homomorphism. Indeed, note first that the target variable of $\unrav{q}{d}$ is mapped via $h$ to the target variable $x$ of $q$, as the former corresponds to the valid path $x$. Second, if a valid path $P$ is labeled by a constant $a$, then $\pathend(P)$ is
labeled by $a$ by definition. In particular, $h(P)=\pathend(P)$ is labeled by $a$. Finally, take and edge $R(P,P')$ in $\unrav{q}{d}$. By definition, we have two cases: either $P'=P,e',\pathend(P')$ or $P=P',e,\pathend(P)$. In the first case, we know that $e'$ is an edge of the form $R(\pathend(P),\pathend(P'))$ in $q$. In particular, $R(h(P),h(P'))$ is an edge in $q$. In the second case, we know that $e$ is an edge of the form $R(\pathend(P),\pathend(P'))$ in $q$. Again, we have that $R(h(P),h(P'))$ is an edge in $q$. We conclude that $h$ is a homomorphism.

\noindent
\underline{$\rhd$ \emph{Conservativeness}}

By Theorem~\ref{thm:cm77}, it suffices to show that in case $q(x)$ is tree-like, we have a homomorphism from $q$ to $\unrav{q}{d}$, where $d$ is the depth of $q$. This follows directly by observing that $q$ is exactly  $\unrav{q}{d}$, modulo renaming of variables. Formally, we can define the ``identity'' homomorphism as follows. For each node $u$ of $q$, we define $h(u)$ to be the unique path in $q$ from the tree root to the node $u$. Note that this is actually a valid path and hence $h$ is well-defined. By construction, $h$ is clearly a homomorphism.

\paragraph{Proof of Proposition~\ref{prop:unravel_prop_opt}}
We next provide the proof of Proposition~\ref{prop:unravel_prop_opt}.

We define a \emph{path} of $q(x)$ as a valid path without the third condition in the definition. That is, in a path we can traverse consecutively the same edge. 

As for valid paths, we denote by $\pathend(P)$ the node at the end of the path $P$. Note that every path $P$ that is not valid define a unique valid path $\pathvalid(P)$ such that $\pathend(P)=\pathend(\pathvalid(P))$. Indeed, whenever we have a subsequence in the path $P$ of form $y,e,z,e,y$ violating the third condition, we can replace it by $y$. By iteratively, applying this simplification, we always obtain a unique valid path $\pathvalid(P)$ with $\pathend(P)=\pathend(\pathvalid(P))$.

Now suppose $q'(x)$ is a tree-like over-approximation of $q(x)$ of depth at most $d$. By Theorem~\ref{thm:cm77}, there exists a homomorphism $h$ from $q'$ to $q$. We shall define a homomorphism $g$ from $q'$ to $\unrav{q}{d}$, which implies $\unrav{q}{d}\subseteq q'$ as required. 

For a node $w$ in $q'$, we define the path $P_w$ in $q$ as follows. Take the unique path from the root $x$ to $w$ in $q'$. As $h$ is a homomorphism, the image of this path via $h$ produces a path in $q$, which we denote $P_w$. Note that $P_w$ is actually a path as $q$ does not have constants. (If $q$ has constants, then $P_w$ could have internal nodes in the sequence that are constants, violating the first condition of valid paths.)

Consider the mapping $g$ from the nodes of $q'$ to the nodes of $\unrav{q}{d}$ defined as follows. For each node $w$ in $q'$, we assign $g(w)=\pathvalid(P_w)$. Note that since the depth of $q'$ is at most $d$, then the length of $P_w$ is at most $d$, and consequently, $\pathvalid(P_w)$ is a valid path of $q$ of length at most $d$. Therefore, $g$ is well-defined.

We claim that $g$ is a homomorphism. Note that, by construction, the target variable of $q'$ is mapped via $g$ to the target variable of $\unrav{q}{d}$. Now, take an edge $R(w,w')$ in $q'$ and suppose $w$ is the parent of $w'$ in $q'$ (the other case is analogous).  
Note that either $\pathvalid(P_{w'})$ extends $\pathvalid(P_{w})$, that is, $\pathvalid(P_{w'}) = \pathvalid(P_{w}), e', z'$, or $\pathvalid(P_{w})$ extends $\pathvalid(P_{w'})$. Indeed, if the last edge $e'$ of $P_{w'}$ is different from the last edge of $\pathvalid(P_{w})$ then $\pathvalid(P_{w'})=\pathvalid(P_{w}), e', z'$. 
Otherwise, if the last edge $e'$ of $P_{w'}$ coincides with the last edge of $\pathvalid(P_{w})$, then $\pathvalid(P_{w'})$ is the subsequence of $\pathvalid(P_{w})$ that ends just before traversing the last edge $e'$ of $\pathvalid(P_{w})$. In particular,  $\pathvalid(P_{w}) = \pathvalid(P_{w'}), e', z$. Suppose $\pathvalid(P_{w'}) = \pathvalid(P_{w}), e', z'$ (the other case is analogous). Note that $e'=R(h(w), h(w'))=R(\pathend(\pathvalid(P_{w})), \pathend(\pathvalid(P_{w'})))$ and $z'=h(w')$. By construction of $\unrav{q}{d}$, we have an edge $R(\pathvalid(P_{w}),\pathvalid(P_{w'}))$ in $\unrav{q}{d}$. 
It follows that $R(g(w), g(w'))=R(\pathvalid(P_{w}),\pathvalid(P_{w'}))$ belongs to $\unrav{q}{d}$ as required.

This concludes the proof. \eop

\paragraph{Proof of Proposition~\ref{prop:unr_depth}}
We verify that $\unrav{q}{d}\subseteq \unrav{q}{d-1}$ using Theorem~\ref{thm:cm77}. This follows from the fact that $\unrav{q}{d-1}$ is a subtree of $\unrav{q}{d}$. Hence we can consider the identity homomorphism that maps each valid path $P$ in $\unrav{q}{d-1}$ to itself.\eop

\section{\baseline: A new probabilistic query answering baseline}\label{sec:app_base_theory_exp}

\paragraph{Possible world semantics.}
A \emph{probabilistic knowledge graph} $(\G,\omega)$ is a knowledge graph $\G$ in which $\omega$ assigns a probability $\omega_R(a,b)$ to each edge $R(a,b)$ in $\G$. The standard
semantics of graph pattern queries (and for any query in any query language for that matter) is defined in terms of the \emph{possible world semantics}.  A \emph{possible world $\W$ of} $\G$ is any subset of edges in $\G$. The probability $\prob_\omega(\W)$ of this world corresponds to
\[
\prod_{R(a,b)\in \W} \omega_R(a,b)\cdot \prod_{R(a,b)\in \G\setminus \W}\bigl(1-\omega_R(a,b)\bigr).\]
That is, $\prob_\omega(\W)$ is the probability of hitting every edge in $\W$ (according to $\omega)$, and missing every edge not in $\W$. 
We refer to the book by \citet{Suciu_prob_book:2011} for further details and background.

Query answering in this context corresponds to computing the probability of each possible world where a particular answer is valid. That is, 
the evaluation of a query $q$ over $(\G,\omega)$ associates to every answer $a$ the probability 
\[
\prob(q,\G,\omega,a):=\sum_{\W\subseteq \G,  a\in q(\W)} \prob_\omega(\W).
\]
This semantics is widely accepted as \emph{the standard} for probabilistic query answering. However, due to the exponential number of possible worlds, evaluating general queries using this method is $\#P$-hard \citep{DBLP:conf/pods/DalviS07a,DalviSuciu_JACM}. 

\paragraph{Hierarchical queries and dissociations.}
To cope with the intractability of computing the possible world semantics, previous work either identified subclasses of queries where evaluation is in PTIME, or constructed approximations.

For tractability, it was shown by \citet{DBLP:conf/pods/DalviS07a} that PTIME evaluation is  possible for so-called self-join-free hierarchical queries. Let $\Neigb(q,x):=\In(q,x)\cup \Out(q,x)$, that is, the set of edges in $q$ adjacent to $x$.
A graph pattern query $q$ is \emph{hierarchical} if for each pair of non-target variables $x$, $y$ in $q$, either $\Neigb(q,x) \subseteq \Neigb(q,y)$, or $\Neigb(q,y) \subseteq \Neigb(q,x)$, or 
$\Neigb(q,x) \cap \Neigb(q,y) = \emptyset$.
If a relation type never occurs more than once in a query, then it is called self-join free.

For approximation, for queries that are not hierarchical, \citet{diss3} developed an approximation technique.
In a nutshell, they showed that every (self-join-free) query
can be \emph{upper bounded} by a finite set of  hierarchical queries, called \emph{dissociations}. In particular, they show that 
\[
\prob(q,\G,\omega,a) \leq \prob(\delta,\G,\omega,a)
\]
for any dissociation $\delta$ of $q$.  \citet{diss3} propose to take $\min_{\delta\in\mathsf{Dis}(q)}\{\prob(\delta,\G,\omega,a)\}$, with $\mathsf{Dis}(q)$ the finite set of dissociations of $q$, as PTIME-computable approximation of 
$\prob(q,\G,\omega,a)$.

\paragraph{\baseline.}
In our new baseline \baseline we use \nbfnets to turn $\G$ into a probabilistic knowledge graph. More precisely, we define for each relation type $R\in\R$, $\omega_R(R(a,b)):=\cP_R(\bm{1}_a)_b$ with
$\cP_R$ is the operator learned by an \nbfnet. We store all $\mathcal O(|\V|^2)$ probabilistic edges.

For query evaluation, \baseline does the following:
\begin{itemize}
    \item If $q(x)$ is tree-like we perform a bottom-up computation, as shown in Algorithm~\ref{alg:propagation}. We here also cover tree-like queries with negation and disjunction.    We remark that, if $q(x)$ is also  hierarchical (also self-join-free, no negation, disjunction), then \baseline actually computes the possible world semantics. Indeed, \baseline coincides  with the PTIME algorithm \citep{DBLP:conf/pods/DalviS07a} for tree-like hierarchical queries
    \item If $q(x)$ is not tree-like but is hierarchical, we evaluate it according to \citet{DBLP:conf/pods/DalviS07a}. In our cases, this only applies to the lollipop query ($\exists$1p2c).
    \item Finally, if $q(x)$ is not-tree like and not hierarchical we compute its dissociations, evaluate these according to their PTIME evaluation algorithm, and take the minimal value, according to the dissociation-based approximation approach of \citet{diss3}.
\end{itemize}

Table~\ref{tab:classification} shows which of the queries used in our experiments satisfy the conditions of being tree-like and/or hierarchical, but without negation or disjunction. 
\begin{table}[t]
\caption{Queries categorized in terms of being hierarchical (H) and/or tree-like. }
\label{tab:classification}
\centering
\begin{tabular}{|c|c|c|c|}
\hline
\textbf{Tree-like+H}
&
\textbf{Tree-like+not H}
&
\textbf{Not tree-like+H}
& 
\textbf{Not tree-like+ not H} \\ \hline
\begin{tabular}[c]{@{}l@{}}
1p,2p,2i,3i,ip,\\ 
$\exists$1p,$\exists$2p,$\exists$2i,$\exists$3i,$\exists$ip\end{tabular} & \begin{tabular}[c]{@{}l@{}}
3p, pi,\\
$\exists$3p,$\exists$pi
\end{tabular} &
$\exists$1p2c 
&  $\exists$3c,$\exists$4c                         \\ \hline
\end{tabular}
\end{table}

The query patterns with disjunction (2u, up, $\exists$2u,  
$\exists$up) and negation (2in, 3in, inp, pni, pin, $\exists$2in, $\exists$3in, $\exists$inp, $\exists$pni, $\exists$pin) are also evaluated in a bottom-up fashion, similarly as \gnnqe, as explained in Algorithm~\ref{alg:propagation}. Dissociations are computed for all non-hierarchical queries. We remark that the maximal number of dissociations was five (for the 4c query).

We note that the \baseline evaluation algorithm for tree-like queries uses $p\mathop{\circled{$\vee$}}q := p + q - pq$ for $p,q\in[0,1]$ to encode disjunction and existential quantification.

\begin{algorithm}[t]
\caption{Evaluation method of $\baseline$ for tree-like queries (extended with negation and disjunction).  The algorithm uses the disjunction operation }\label{alg:propagation}
\begin{algorithmic}[1]
\Require{tree-like query $q(x)$, probabilistic knowledge graph $(\G,\omega)$.}
\Ensure{$\mathsf{score}_\omega(q,\G)(e)$ for entity $e$}

\If {$q(x)$ is of the form  $R(a,x)$ (resp. $R(x,a)$)}
\State \Return $\omega_R(a,e)$ (resp.  $\omega_{R}(e,a)$).
\EndIf
\If {$q(x)$ is of the form  $R(y,x)$ (resp. $R(y,a)$)}
\State \Return $\circled{$\bigvee$}_{a\in\V}\omega_R(a,e)$  (resp.  $\circled{$\bigvee$}_{a\in\V}\omega_R(e,a)$).
\EndIf
\If {$q(x)$ can be decomposed in two
$q_1(x)$ and $q_2(x)$ with disjoint variable nodes (except for the target variable $x$}
\State \Return $\mathsf{score}_\omega(q_1,\G)(e)\cdot \mathsf{score}_\omega(q_1,\G)(e)$.
\EndIf
\If {$q(x)$ can be decomposed in a query $q_1(y)$ not containing $x$ and and edge $R(y,x)$ (resp. $R(x,y)$)}
\State \Return 
$\circled{$\bigvee$}_{a\in\V}\mathsf{score}_\omega(q_1,\G)(a)\omega_R(a,e)$ 
(resp. $\circled{$\bigvee$}_{a\in\V}\mathsf{score}_\omega(q_1,\G)(a)\omega_R(e,a)$. 
\EndIf
\If {$q(x)$ is of the form $\neg q_1(x)$}
\State \Return $1-\mathsf{score}_\omega(q_1,\G)(e)$.
\EndIf
\If {$q(x)$ is of the form $q_1(x)\vee q_2(x)$}
\State \Return $\mathsf{score}_\omega(q_1,\G)(e)\mathop{\circled{$\vee$}}\mathsf{score}_\omega(q_2,\G)(e)$.
\EndIf
\end{algorithmic}
\end{algorithm}

\section{Experimental details}

\subsection{Implementation and training details} \label{sec:app-training-implementation}

\paragraph{\nbfnet.} Table \ref{tab:learning-hyper-nbfnet} shows the training hyperparameters used to train model \nbfnet for \baseline. The parameters are kept as in the original paper \citep{zhu2021bellman}. The best checkpoint for the model is selected based on the performance in the validation set, using MRR as the selection criteria. 

\begin{table}[t]
\centering
\caption{Learning hyperparameters of \nbfnet for $\baseline$ framework for FB15k237. The hyperparameters are kept as in the original \nbfnet paper.} 
\label{tab:learning-hyper-nbfnet}
\resizebox{0.48\textwidth}{!}{%
\begin{tabular}{llll}
&\textbf{Hyperparameter} &  & \textbf{FB15k-237} \\ \hline
\multirow{2}{*}{\bf GNN}              & \#layer              &  & 6                              \\
& hidden dim.               &  & 32                         \\ \hline  
\multirow{2}{*}{\bf MLP}              & \#layer              &  & 2                              \\
& hidden dim.               &  & 64                         \\ \hline 
\multirow{2}{*}{\bf Batch}              & \#positive              &  & 256                              \\
& \#negative/\#positive               &  & 32                         \\ \hline 
\multirow{4}{*}{\bf Learning}              & optimizer              &  & Adam                              \\
& learning rate               &  & 5e-3                         \\ 
& \#epoch               &  & 20                         \\ 
& adv. temperature               &  & 0.5                         \\ \hline 
\end{tabular}
}
\end{table}

\paragraph{\gnnqe.}
Table \ref{tab:learning-hyper} shows the training hyperparameters used to train model \gnnqe within our framework. The parameters are kept as in the original paper \citep{zhu2022neural}. The best checkpoint for the model is selected based on the performance in the validation set, using MRR as the selection criteria.
\begin{table}[t]
\centering
\caption{Learning hyperparameters of \gnnqe in $\qta$ framework for FB15k237, FB15k and NELL.} 
\label{tab:learning-hyper}
\resizebox{0.55\textwidth}{!}{%
\begin{tabular}{lllll}
\textbf{Hyperparameter} &  & \textbf{FB15k-237} & \textbf{FB15k} & \textbf{NELL} \\ \hline
Batch size              &  & 24                 & 24             & 6             \\
Optimizer               &  & Adam               & Adam           & Adam          \\
Learning rate           &  & 5e-3               & 5e-3           & 5e-3          \\
Adv. temperature        &  & 0.2                & 0.2            & 0.2           \\
\#Batches               &  & 10.000             & 10.000         & 18.000       
\end{tabular}%
}
\end{table}

To train and evaluate we used a GPU Titan RTX 24.2 GB. We also provide the configuration files (.yaml) in the source code.  

\subsection{Message Passing Query Embedding}
\label{sec:app-mpqe-details}
Our experiments with \mpqe are based on \textsf{StarQE}~\citep{alivanistos2021query}, a method for answering tree-like queries on \emph{hyper-relational} knowledge graphs, where edges can have \emph{qualifiers} that provide additional information about a triple. When ignoring qualifiers, \textsf{StarQE} can be seen as an adaptation of \mpqe that employs a \textsf{CompGCN} model~\citep{VashishthSNT20} for encoding queries. We rely on the implementation of \textsf{StarQE} available online.\footnote{\url{https://github.com/DimitrisAlivas/StarQE}}

\subsection{The effect of training with general tree-like queries} \label{appendix:anchor+unanchored} 

One characteristic of our framework is that we extend current neuro-symbolic methods to deal with tree-like queries with and without anchors. In Tables \ref{tab:all-results-og} and \ref{tab:all-results-sa} we report metrics over anchored and unanchored tree-like queries for the same model when including the later in the training set. 

For anchored tree-like queries we compare against \gnnqe (see Table \ref{tab:all-results-og}). The results are taken from the original paper
\renewcommand{\arraystretch}{1.12}
\begin{table}[t]
\caption{Results of evaluation of \gnnqe and \qta over original test queries (anchored tree-like) over FB15k-237, FB15k and NELL995. Metrics of \gnnqe are taken from its original paper. }
\label{tab:all-results-og}
\vspace{0.2cm}
\resizebox{\textwidth}{!}{%
\begin{tabular}{llllllllllllllll}
\bf Metric                     & \bf Model            & 
\bf 1p & \bf 2p & \bf 3p &\bf  2i &\bf  3i & \bf ip & \bf pi &\bf  2in &\bf  3in & \bf inp &\bf  pin & \bf pni &\bf  2u &\bf  up \\ \hline
\multicolumn{16}{|c|}{\textbf{FB15k-237}} \\ \hline
\multirow{2}{*}{hits@1}    & \gnnqe           &  0.328  & \bf 0.082   & \bf 0.065   & \bf 0.277   & 0.446   &  0.123  & 0.224   &  \bf 0.041   &  \bf 0.081  &   \bf 0.041  &  \bf 0.025   &  \bf 0.027   & \bf 0.098   &  \bf 0.076  \\

& $\qta$ & \bf 0.343  &  0.055  &  0.018  &  \bf 0.277  &  \bf 0.479  &  \bf 0.138  &  \bf 0.227  &  0.016   &  0.074   &  0.037   &  0.017   &  0.013   &  0.087  &  0.058  \\ \hline
\multirow{2}{*}{mrr}       & \gnnqe           &     0.428        &     \bf0.147        &      \bf0.118       &    \bf0.383         &      0.541       &       0.189      &    \bf0.311        &       \bf0.100       &    \bf0.168          &      \bf0.093        &     \bf 0.072         &    \bf 0.078         &     \bf0.162        &     \bf0.134       \\

& $\qta$ &  \bf 0.431  &  0.093  &  0.071  &  0.364  &  \bf 0.595  &  \bf 0.190  &  0.300  &  0.048   &  0.128   &   0.085   &  0.045   &   0.051  &  0.134  &  0.095  \\ \hline
\multicolumn{16}{|c|}{\textbf{FB15k}}                                                                                    \\ \hline
\multirow{2}{*}{hits@1}    & \gnnqe           &  \bf 0.861  &  0.635  &  0.525  &  0.748  &  \bf 0.801  &  \bf 0.651  &  0.636  & \bf 0.354    &  \bf 0.331   & \bf 0.338    &  0.186   &  0.218   & 0.671   &  0.530  \\

& $\qta$ &  0.831  &  \bf 0.643  &  \bf 0.533  &  \bf 0.759  &  0.791  &  0.566  &  \bf 0.668  &   0.308  &  0.297   &  0.315   &  \bf 0.189   &  \bf 0.219   &  \bf 0.704  &  \bf 0.540  \\
& \mpqe & 0.258 & 0.061 & 0.060 & 0.094 & 0.121 & 0.081 & 0.096 & - & - & - & - & - & - & - \\ \hline
\multirow{2}{*}{mrr}       & \gnnqe           &      \bf0.885      &       \bf0.693      &     \bf 0.587        &    0.797         &      \bf0.835       &     \bf0.704        &      0.699       &      \bf0.447        &      0.417        &      \bf0.420        &     0.301         &     \bf0.343         &     0.741        &     \bf0.610        \\
& $\qta$ &     0.855        &     0.688        &      \bf 0.587       &       \bf0.801      &       0.833      &        0.620     &     \bf0.720        &      0.430        &      \bf0.418        &      0.403        &      \bf0.302        &     0.340         &    \bf0.747         &  0.600   \\ 
& \mpqe & 0.362 & 0.107 & 0.100 & 0.153 & 0.198 & 0.138 & 0.145 & - & - & - & - & - & - & -  \\ \hline
\multicolumn{16}{|c|}{\textbf{NELL995}}                 \\ \hline
\multirow{2}{*}{hits@1}    & \gnnqe           &  \bf 0.861  &  0.635  &  0.525  &  0.748  &  \bf 0.801  &  \bf 0.651  &  \bf 0.636  & \bf 0.354    &  \bf 0.331   & \bf 0.338    &  0.186   &  0.218   & 0.671   &  0.530  \\
& $\qta$ &  0.831  &  \bf 0.643  &  \bf 0.533  &  \bf 0.759  &  0.791  &  0.566  &  0.668  &   0.308  &  0.297   &  0.315   &  \bf 0.189   &  \bf 0.219   &  \bf 0.704  &  \bf 0.540   \\ 
& \mpqe & 0.191 & 0.032 & 0.046 & 0.066 & 0.114 & 0.031 & 0.056 & - & - & - & - & - & - & - \\ \hline
\multirow{2}{*}{mrr}       & \gnnqe       & \bf 0.533       & \bf 0.189       & \bf 0.149       & \bf 0.424       & \bf 0.525       & \bf 0.189       & \bf 0.308       & \bf 0.159        & 0.126        & 0.099        & \bf 0.146        & \bf 0.114        & 0.063       & 0.063       \\
& $\qta$    & 0.479       & 0.160       & 0.119       & 0.378       & 0.484       & 0.140       & 0.269       & 0.086        & 0.128        & \bf 0.111        & 0.055        & 0.053       & \bf 0.141       & \bf 0.107 \\ 
& \mpqe & 0.270 & 0.061 & 0.075 & 0.132 & 0.198 & 0.060 & 0.097 & - & - & - & - & - & - & - \\ \hline
\end{tabular}%
}
\end{table}. We note in general that \unr tends to inherit the peformance of \gnnqe to deal with anchored queries, with some exceptions where the performance slightly decay or increase. 

For unanchored tree-like queries, in Table \ref{tab:all-results-sa} we compare \unr with a-\unr that has the same architecture as \unr but that is trained only using anchored tree-like queries. As expected, when unanchored tree-like queries are included in the training set, the performance on such queries tend to increase. 
\renewcommand{\arraystretch}{1.1}
\begin{table}[t]
\caption{Comparison of results of evaluation of \qta over test set of unanchored queries when including (\qta) or not including (a-\qta) unanchored queries in the training phase. }
\label{tab:all-results-sa}
\vspace{0.2cm}
\resizebox{\textwidth}{!}{%
\begin{tabular}{llllllllllllllll}
\bf Metric                     & \bf Model            & \bf 1p & \bf 2p & \bf 3p &\bf  2i &\bf  3i & \bf ip & \bf pi &\bf  2in &\bf  3in & \bf inp &\bf  pin & \bf pni &\bf  2u &\bf  up \\ \hline
\multicolumn{16}{|c|}{\textbf{FB15k-237}}                                                                                \\ \hline
\multirow{2}{*}{hits@1}    & a-\qta &  0.001  &  0.042  &  0.028  &  0.119  & 0.189   &  \bf 0.060  & \bf 0.104   & 0.005    &  \bf 0.057   &   0.025  &  0.020   &   0.001  &  0.018  &    0.016   \\ 
& \qta &  \bf 0.028  &  \bf 0.046  &  \bf 0.057  &  \bf 0.123  &  \bf 0.203  &  0.030  &  0.090  &  \bf0.011   & 0.055    &  \bf 0.028   &  \bf 0.019   &  \bf 0.005   &  \bf 0.030  &    \bf 0.023   \\ 
& \mpqe & 0.012 & 0.041 & 0.035 & 0.021 & 0.027 & 0.030 & 0.028   & - & - & - & - & - & - & - \\ \hline
\multirow{2}{*}{mrr}       & a-\qta           &  0.012  &  0.081  &  0.063  &  0.206 &  0.286  & \bf 0.108   & \bf 0.165   &   0.024  &  \bf 0.108   &  0.059   &  0.044   &  0.019   &  0.040  &  0.049  \\
& \qta &  \bf 0.055  & \bf 0.088   &  \bf 0.079  &  \bf 
 0.212  &  \bf 0.292  &  0.076  &  0.151  &   \bf 0.029  &  0.107   &  \bf 0.061   &   \bf 0.045  &  \bf 0.027   & \bf 0.065   &   \bf 0.053 \\ 
 & \mpqe & 0.029 & 0.069 & 0.061 & 0.041 & 0.049 & 0.052 & 0.046 & - & - & - & - & - & - & - \\ \hline
\multicolumn{16}{|c|}{\textbf{FB15k}}                                                                                    \\ \hline
\multirow{2}{*}{hits@1}    & a-\qta           &  0.075  & 0.315   & 0.387   & 0.552   & 0.620   &  0.511  & 0.470   &  0.054   &  0.178   &   0.162  &  0.074   &  0.050   & 0.084   &  0.347  \\

& $\qta$   &  \bf 0.171  & \bf 0.353   & \bf 0.426   & \bf 0.628   & \bf 0.661   &  \bf 0.523  & \bf 0.511   &  \bf 0.103   &  \bf 0.209   &   \bf 0.181  &  \bf 0.094   &  \bf 0.081   & \bf 0.209   &  \bf 0.373  \\ 
& \mpqe & 0.083 & 0.057 & 0.047 & 0.053 & 0.069 & 0.083 & 0.041 & - & - & - & - & - & - & - \\ \hline
\multirow{2}{*}{mrr}       & a-\qta           &  0.112    &     0.387        &   0.457          &     \bf 0.685        &     0.686        &     0.571        &      0.533       &     0.109        &     0.283         &     0.240         &     0.152         &        0.102      &      0.131        &    0.421 \\

& \qta & \bf0.228        &     \bf0.434        &      \bf0.496       &      0.613       &      \bf 0.718       &      \bf 0.579       &     \bf0.572        &   \bf 0.190           &      \bf0.325        &   \bf0.269           &      \bf0.189        &     \bf0.158         &       \bf0.292      &    \bf0.455 \\ 
& \mpqe & 0.148 & 0.092 & 0.084 & 0.094 & 0.115 & 0.126 & 0.074 & - & - & - & - & - & - & -\\ \hline
\multicolumn{16}{|c|}{\textbf{NELL}}              \\ \hline
\multirow{2}{*}{hits@1}    & a-\qta &  0.004  & 0.022   & 0.033   &  0.130  & 0.183   &  0.057  &  0.082  & 0.066    &  0.021   &  0.025   &  0.004   &  0.004   & 0.005   &  \bf 0.017    \\

& $\qta$ &  \bf 0.010  & 0.029   & 0.038   &  \bf 0.142  & \bf 0.190   &  0,056  &  \bf 0.084  & \bf 0.097    &  \bf 0.022 &  \bf 0.027   &  \bf 0.006   &  \bf 0.006   & \bf 0.012   &  0.012  \\ 
& \mpqe & 0.008 & \bf 0.034 & \bf 0.050 & 0.034 & 0.063 & \bf 0.061 & 0.038 & - & - & - & - & - & - & - \\ \hline
\multirow{2}{*}{mrr}       & a-\qta           &  0.007  & 0.052   & 0.069   &  0.211  & 0.276   &  \bf 0.110  &  \bf 0.139  & 0.023    &  0.075   &  0.059   &  0.025   &  0.018   & 0.012   &  0.043  \\

& $\qta$ &  \bf 0.028  & \bf 0.067   & \bf 0.075   &  \bf 0.228  & \bf 0.290   &  0.107  &  0.137  & \bf 0.034    &  \bf 0.080   &  \bf 0.066   &  \bf 0.033   &  \bf 0.024   & \bf 0.033   &  \bf 0.047  \\ 
& \mpqe & 0.023 & 0.056 & 0.075 & \bf 0.075 & 0.125 & 0.082 & 0.073 & - & - & - & - & - & - & - \\ \hline
\end{tabular}%
}
\end{table}

\subsection{Results of cyclic queries for FB15k and NELL995} \label{sec:app-cyclic-fb15k-nell}

Table \ref{tab:cyclic-other-datasets} presents the results for cyclic queries in datasets FB15k and NELL995. 

This results further show that \unr is a performant framework to deal with complex queries. 

\renewcommand{\arraystretch}{1.1}
\begin{table}[t]
\centering
\caption{Results for cyclic queries considering the best unravelings in datasets FB15k and NELL995. }
\label{tab:cyclic-other-datasets}
\vspace{0.2cm}
\resizebox{0.8\textwidth}{!}{%
\begin{tabular}{llllllll}
\bf Metric                     &     \bf  Model      & $\pmb\exists$\textbf{1p2c} & $\pmb\exists$\textbf{3c} & $\pmb\exists$\textbf{4c}    \\ \hline
\multicolumn{5}{c}{\textbf{Cyclic queries FB15k}}  \\ \hline

\multirow{2}{*}{mrr}& \qta &  0.098  & \bf 0.297   & 0.159          \\
& \mpqe &  \bf 0.120 & 0.116 & 0.000         \\ \hline
\multirow{2}{*}{hits@1} & \qta  &  0.057  & \bf 0.092   & 0.057          \\ 
&  \mpqe & \bf 0.064 & 0.063 & 0.000    \\ \hline
\multirow{2}{*}{hits@10} & \qta &  \bf 0.227  & \bf 0.375   & 0.302          \\
& \mpqe & 0.225 & 0.218 & 0.000        \\ \hline
\end{tabular}%
\quad
\begin{tabular}{llllllll}
\bf Metric                     &    \bf  Model       & $\pmb\exists$\textbf{1p2c} & $\pmb\exists$\textbf{3c} & $\pmb\exists$\textbf{4c}    \\ \hline
\multicolumn{5}{c}{\textbf{Cyclic queries NELL995}}  \\ \hline

\multirow{2}{*}{mrr} & \qta &  0.146  & \bf 0.211   & 0.122          \\ 
& \mpqe & \bf 0.150 & 0.142 & \bf 0.180       \\ \hline
\multirow{2}{*}{hits@1} & \qta &  0.065  & \bf 0.114   & 0.067          \\
& \mpqe & \bf 0.074 & 0.060 & \bf 0.098      \\ \hline
\multirow{2}{*}{hits@10} & \qta &  \bf 0.086  & \bf 0.419   & \bf 0.257          \\
& \mpqe &   0.074 & 0.060 & 0.098       \\ \hline
\end{tabular}%
     }
\end{table}

\subsection{Further results for FB15k-237} 
\label{appendix:more-results}

\paragraph{Unravelings depth. } While the unraveling depth does not impact much on ranking metrics such as mrr and hits@1, when we look at more \emph{global} metrics, such as precision and recall, some tendencies appear. We can see in Table \ref{tab:classification-metrics-cyclic} the results of \unr for the triangle and square queries when considering different depths and different classification thresholds

\renewcommand{\arraystretch}{1.1}
\begin{table}[t]
\caption{Precision and recall (\%) for $\qta$ at classification thresholds 0.3, 0.5, 0.7 for the triangle (unravellings depth 2 to 8) and square (unravellings depth 3 to 9) queries. For recall we consider both easy and hard answers (see Recall in table) and only hard answers (Hard-Recall) }
\label{tab:classification-metrics-cyclic}
\vspace{0.2cm}
\resizebox{\textwidth}{!}{
\begin{tabular}{c|l|ccccccc|ccccccc}
\textbf{T} & \textbf{Metric}       & \multicolumn{7}{c}{\textbf{Triangle query (3c)}} & \multicolumn{7}{c}{\textbf{Square Query (4c)}} \\
&                       & 2     & 3     & 4    & 5    & 6    & 7    & 8    & 3    & 4    & 5    & 6    & 7    & 8    & 9    \\ \hline
\multirow{3}{*}{0.3}              & Precision             &    34.9    &    50.7   &    53.4  & 54.8    &   {\bf 55.2}   &   53.1    &  52.2    
&   58.2   &   {\bf 61.6}   &  60.8    &   55.4   &    51.4  &  47.4    &   43.4   \\
& Recall                &    {\bf 90.8}   &   82.1    &   73.2   &   66.9   &   61.7   &   58.0   &   55.3   &   \bf 75.5   &  60.7    &   46.5   &  36.7    &  30.6    &  26.8    &  24.9    \\
& Hard-Recall &   \bf 66.3    &   46.8    &   36.4   &   30.7   &   28.0   &   25.7   &   24.5   &   \bf 33.8   &  25.5    &  18.3    &  12.2    &   8.9  &  7.1    &   6.6   \\ \hline
\multirow{3}{*}{0.5}              & Precision             &   46.3    &   54.0    &   \bf 55.1   &   54.8   &    51.9  &   50.0   &   49.9   &   \bf 48.7   &   47.6   &  42.6    &   37.6   &   34.8   &   32.6   &   32.3   \\
& Recall                &    \bf 73.9   &   57.8    &  50.3    &   45.7   &   41.5   &   38.9   &   37.8   &   \bf 38.4   &  22.6    &   17.5   &   15.2   &  14.0    &   13.4   &   13.0   \\
& Hard-Recall  &    \bf 43.2   &   26.0    &   18.8   &   16.1   &   14.1   &   13.4   &   12.9   &   \bf 13.1   &  5.6    &   3.0   &  1.8    &   1.6   &  1.1    &  1.0    \\ \hline
\multirow{3}{*}{0.7}              & Precision             &   38.4    &   42.3    &   44.1   &   \bf 45.0   &  43.9    &   44.4   &   43.9          &   \bf 30.5   &   23.8   &   23.8   &   23.1   &  22.8    &  23.0 &  23.0 \\
 & Recall                &    \bf 45.9   &   31.2    &   27.7   &   26.6   &   25.2   &   24.8   &   24.7   &  \bf 8.4    &   6.4   &   6.3   &   6.4   &   6.4   &  6.3    &   6.3   \\
& Hard-Recall &   \bf 25.2    &   12.6    &   9.9       &   8.5   &    7.6  &   7.7   &   7.4   &   \bf 0.7   &  0.3    &   0.4   &   0.3   &   0.1   &   0.1 &   0.0   
\end{tabular}
}
\end{table}

\begin{figure}[t]
\label{fig:unrav-depths}
\centering
\begin{minipage}{.48\textwidth}
  \centering
  \includegraphics[width=.9\linewidth]{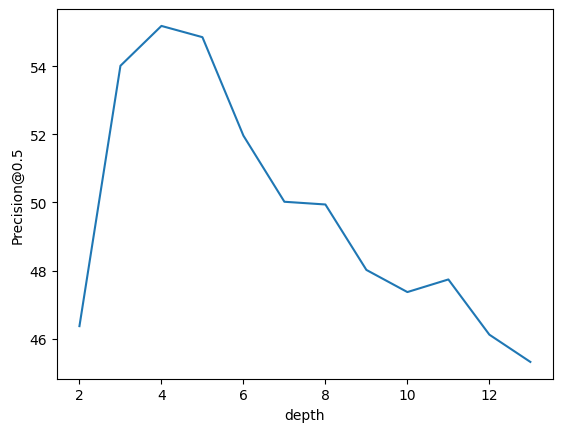}
  \captionof{figure}{Precision using 0.5 classification threshold for different depths of unravelings for the triangle query (FB15k-237).}
  \label{fig:triangle-precision0.5}
\end{minipage}%
\quad
\begin{minipage}{.48\textwidth}
  \centering
  \includegraphics[width=.9\linewidth]{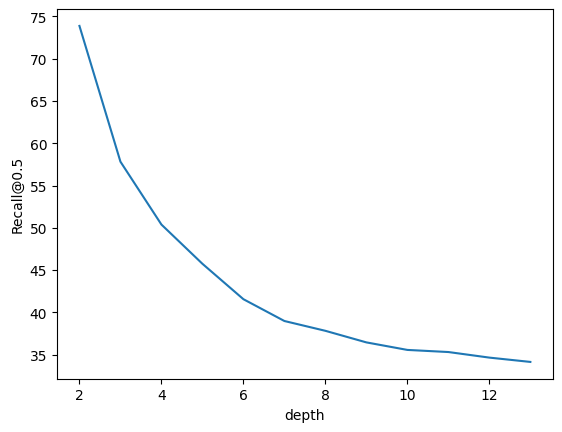}
  \captionof{figure}{Recall using 0.5 classification threshold for different depths of unravelings for the triangle query (FB15k-237).}
  \label{fig:triangle-recall0.5}
\end{minipage}
\end{figure}

It can be noted that the best performance occurs at different depths for different thresholds. Moreover, each metric follows a different tendency:

\begin{itemize}
    \item \textbf{Precision}: Looking at precision against the unravelings depth, we note that it starts growing, it peaks and then starts decaying. Figure \ref{fig:triangle-precision0.5} shows this for the triangle query with classification threshold 0.5. In this case, the peak occurs at depth $d = 0.4$
    \item \textbf{Recall}: In most cases, recall tends to peak with shorter depths and then starts decaying. Figure \ref{fig:triangle-recall0.5} shows this for the triangle query with classification threshold 0.5 
    \item \textbf{Hard-Recall}: Recall, when only considering \emph{hard answers}, maintains the same tendency as the general recall: it peaks with shorter unravelings and then starts decaying.
\end{itemize}

\subsection{Query sets }\label{appendix:query-set}
In this paper we present two new query sets for knowledge graphs FB15k, FB15k-237 and NELL995. 
\paragraph{General tree-like queries.} We built a new set of general tree-like queries, these are tree-like queries that include existential leaves. We generate this query set by removing anchors for the queries in the original betaE query set. To denote existential quantified leaves, we use $-5$. For example. take query $Q(x) \gets R(u,x)$. In betaE format this query would be $(u, R)$. Let's say we remove the anchor $u$, then the query would be $Q(x) \gets \exists y. R(y,x)$ and, in betaE format, $(-5, R)$.

Complete statistics for these queries can be found in Table \ref{tab:queries-stats-sa}
\begin{table}[t]
\caption{Statistic of unanchored query set for each dataset: FB15k-237, FB15k and NELL. }
\resizebox{\textwidth}{!}{%
\begin{tabular}{llllllllllllllll}
\label{tab:queries-stats-sa}
& \textbf{Split} & $\pmb\exists$\textbf{1p} & $\pmb\exists$\textbf{2p} & $
\pmb\exists$\textbf{3p} & $\pmb\exists$\textbf{2i} & $\pmb\exists$\textbf{3i} & $\pmb\exists$\textbf{ip} & $\pmb\exists$\textbf{pi} & $\pmb\exists$\textbf{2in} &$\pmb\exists$\textbf{3in} & $\pmb\exists$\textbf{inp} & $\pmb\exists$\textbf{pin} & $\pmb\exists$\textbf{pni} & $\pmb\exists$\textbf{2u} & $\pmb\exists$\textbf{up} \\\hline
\multirow{3}{*}{\textbf{FB15k-237}} & train &474&13139&62826&117688&147721&11644&12790&5719&12286&13358&-&-&-&-               \\
& valid&288&2514&4213&3763&3368&2272&4255&3330&4080&1970&4135&4396&2905&1613             \\
& test&295&2475&4234&3792&3471&2259&4219&3272&3941&1975&3903&4312&2901&1604             \\ \hline 
\multirow{3}{*}{\textbf{FB15k}}     & train&2690    &   51378   &   172943  &   231273  &   271760  &   22250   &   23759   &   11016  &    23337   &   25222   &-&-&-&-              \\
& valid&1182&5246&7481&6499&5072&3592&6908&5825&6506&3173&6620&7093&4916&2559             \\
& test&1240&5302&7433&6527&5252&3558&6902&5815&6442&2979&6306&7125&4906&2565             \\ \hline 
\multirow{3}{*}{\textbf{NELL}}      & train&400&8713&35045&50076&37010&4458&3015&1035&4218&5026&-&-&-&-             \\
& valid&346&2118&3239&2731&2342&1721&3193&2643&2884&1410&2772&3421&2428&1249             \\
& test&342&2050&3192&1596&897&474&1725&1568&1097&373&1542&1841&922&241             \\ \hline            
\end{tabular}%
}
\end{table}

\paragraph{Cyclic queries.} Table \ref{tab:queries-cyclic-stats} show statistics for the cyclic queries test set. 

\begin{table}[t]
\caption{Statistic of cyclic query set for each dataset: FB15k-237, FB15k and NELL. }
\label{tab:queries-cyclic-stats}
\centering
\resizebox{0.32\textwidth}{!}{
\begin{tabular}{llll}
\textbf{Dataset}   & \textbf{2c} & \textbf{3c} & \textbf{4c} \\ \hline
\textbf{FB15k-237} &     1452        & 1047        & 909         \\ 
\textbf{FB15k}     &      1500       &      4364       &     260        \\
\textbf{NELL995}   &     1500      &    623         &  1952          
\end{tabular}
 }
\end{table}

\end{document}